\pdfoutput=1
\documentclass[11pt]{article}

\usepackage[final]{EMNLP2023}

\usepackage{times}
\usepackage{latexsym}
\usepackage{multirow}
\usepackage{tabularx}
\usepackage{graphicx}
\usepackage{booktabs}

\usepackage[T1]{fontenc}
\usepackage[utf8]{inputenc}
\usepackage{algorithm,algpseudocode}
\usepackage{microtype}

\usepackage{inconsolata}
\usepackage[mathscr]{euscript}
\usepackage{subcaption}
\usepackage{array}
\usepackage{footnote}
\usepackage{dingbat}
\usepackage{svg}
\usepackage{amsmath}
\usepackage{amsfonts}
\usepackage{mathtools}
\usepackage[normalem]{ulem}
\usepackage{bchart}
\usepackage{pgfplots}
\usepackage{hyperref}
\usepackage{amssymb}
\usepackage{pifont}
\usepackage{adjustbox}
\usepackage{xr}
\usepackage{collcell}
\usepackage{colortbl}
\usepackage{todonotes}
\usepackage{appendix}
\usepackage{lipsum}                     % Dummytext
\usepackage{xargs}                      % Use more than one optional parameter in a new commands
\usepackage{mathrsfs}
\usepackage{nccmath}
\usepackage{stmaryrd}
\usepackage[inline]{enumitem}
\newcommand{\mbert}{M-BERT}
\newcommand{\xlm}{XLM}
\newcommand{\xlmr}{XLM-R}
\newcommand{\ensquad}{SQUAD{\mbox{\tiny{EN}}}}
\newcommand{\finetune}{Fine-tune}
\newcommand{\sbert}{S-BERT}
\newcommand{\maml}{MAML}
\newcommand{\teachmaml}{T-MAML}
\newcommand{\studmaml}{S-MAML}
\newcommand{\alignmaml}{MAML-Align}
\newcommand{\setset}{\mathcal{D}}

\newcommand{\lareqa}{LAReQA}
\newcommand{\xquadr}{XQuAD-R}
\newcommand{\stsbm}{STSB{\mbox{\tiny{Multi}}}}
\newcommand{\stsben}{STSB{\mbox{\tiny{EN}}}}

\newcommand{\metatraindata}{\setset_{\mbox{\tiny{meta-train}}}}
\newcommand{\metavaliddata}{\setset_{\mbox{\tiny{meta-valid}}}}
\newcommand{\metatestdata}{\setset_{\mbox{\tiny{meta-test}}}}

\newcommand{\metatrainspt}{\setset^{\mbox{\tiny{train}}}_{\mbox{\tiny{support}}}}
\newcommand{\metatrainqry}{\setset^{\mbox{\tiny{train}}}_{\mbox{\tiny{query}}}}

\newcommand{\metatestspt}{\setset^{\mbox{\tiny{test}}}_{\mbox{\tiny{support}}}}
\newcommand{\metatestqry}{\setset^{\mbox{\tiny{test}}}_{\mbox{\tiny{query}}}}

\newcommand{\dev}{Dev}
\newcommand{\train}{Train}
\newcommand{\test}{Test}

\newcommand{\mono}{mono}
\newcommand{\bil}{bi}
\newcommand{\multi}{multi}

\newcommand{\monomono}{\mono$\rightarrow$\mono}
\newcommand{\monobil}{\mono$\rightarrow$\bil}
\newcommand{\monomulti}{\mono$\rightarrow$\multi}

\newcommand{\bilmulti}{\bil$\rightarrow$\multi}
\newcommand{\mixt}{mixt}
\newcommand{\trans}{trans}
\newcommand{\monobilmulti}{\mono$\rightarrow$\bil$\rightarrow$\multi}

\newcommand{\translatetrain}{T-Train}

\newcommand{\labse}{LaBSE}
\newcommand{\laser}{LASER}

\newcommand{\xprime}[1]{#1\prime}

\newcommand{\lng}{\boldsymbol\ell}

\newcommand{\alllang}{\mathscr{L}}

\newcommand{\allqueries}{\mathscr{Q}}
\newcommand{\allcontent}{\mathscr{R}}

\newcommand{\map}{mAP@20}
\newcommand{\pearcorr}{Pearson’s r × 100}

\definecolor{bblue}{HTML}{4F81BD}
\definecolor{rred}{HTML}{C0504D}
\definecolor{ggreen}{HTML}{9BBB59}
\definecolor{ppurple}{HTML}{9F4C7C}

\newcommand{\colorann}[3]{\textcolor{#1}{${}^{#2}[$#3$]$}}

\newcommand{\meryem}[1]{\colorann{green}{--meryem}{#1}}

\newcommandx{\unsure}[2][1=]{\todo[linecolor=red,backgroundcolor=red!25,bordercolor=red,#1]{#2}}
\newcommandx{\change}[2][1=]{\todo[linecolor=blue,backgroundcolor=blue!25,bordercolor=blue,#1]{#2}}
\newcommandx{\info}[2][1=]{\todo[linecolor=OliveGreen,backgroundcolor=OliveGreen!25,bordercolor=OliveGreen,#1]{#2}}
\newcommandx{\improvement}[2][1=]{\todo[linecolor=Plum,backgroundcolor=Plum!25,bordercolor=Plum,#1]{#2}}
\newcommandx{\thiswillnotshow}[2][1=]{\todo[disable,#1]{#2}}

\author{
Meryem M\textquotesingle hamdi$^{1}$\thanks{\hspace{1.5mm}Work was conducted while the first author was a research scientist intern at Adobe.} , Jonathan May$^{1}$, Franck Dernoncourt$^{2}$,\\ \textbf{Trung Bui$^{2}$, and Seunghyun Yoon$^{2}$}\\
$^{1}$Information Sciences Institute, University of Southern California\\% (USC), Los Angeles, USA \\
{\{\tt meryem}, and {\tt jonmay}\}@isi.edu \\
$^{2}$Adobe Research \\%, San Jose, CA, USA\\
{\{\tt dernonco}, {\tt bui}, and {\tt syoon}\}@adobe.com \\
\\}

\title{Multilingual Sentence-Level Semantic Search \\using Meta-Distillation Learning}

\begin{document}
\maketitle

\begin{abstract}
% Given the sheer amount of work done on monolingual retrieval/search and its extension to multilingual scenarios using machine translation, we focus in this paper on developing an efficient solution to multilingual semantic search with less reliance on machine translation. Machine translation solutions are often not efficient and not feasible as the retrieved content in the search can be from multiple languages and it is hard to predict which language combinations are involved in real-time. 
% We aim to reduce the gap between different languages used in the query and to be retrieved sentences. We aim to democratize multilingual semantic search for low-resource evaluation and ad-hoc semantic search. For that purpose, we propose a meta-learning approach based on MAML and show its gain compared to standard fine-tuning. We compare to different external baselines based on machine translation and knowledge distillation and propose an alignment approach based on meta-distillation learning to align monolingual and bilingual semantic search to multilingual semantic search. 

% With the advent of Transformer-based models such as M-BERT, XLM-R and their corresponding sentence encoders, neural retrieval models have overtaken keyword-based approaches, making it easier to find semantically relevant content more efficiently.However, unlike monolingual and bilingual retrieval tasks, Multilingual semantic search is requires at the same time a deeper understanding of the query and a stronger alignment between languages is still under-explored.

Multilingual semantic search is the task of retrieving relevant contents to a query expressed in different language combinations. This requires a better semantic understanding of the user's intent and its contextual meaning. Multilingual semantic search is less explored and more challenging than its monolingual or bilingual counterparts, due to the lack of multilingual parallel resources for this task and the need to circumvent ``language bias''. In this work, we propose an alignment approach: \alignmaml{},\footnote{We will release our code repository in the camera-ready version.} specifically for low-resource scenarios. Our approach leverages meta-distillation learning based on MAML, an optimization-based Model-Agnostic Meta-Learner. \alignmaml{} distills knowledge from a \textbf{T}eacher meta-transfer model \teachmaml{}, specialized in transferring from monolingual to bilingual semantic search, to a \textbf{S}tudent model \studmaml{}, which meta-transfers from bilingual to multilingual semantic search. To the best of our knowledge, we are the first to extend meta-distillation to a multilingual search application. Our empirical results show that on top of a strong baseline based on sentence transformers, our meta-distillation approach boosts the gains provided by MAML and significantly outperforms naive fine-tuning methods. Furthermore, multilingual meta-distillation learning improves generalization even to unseen languages.  

\end{abstract}
\section{Introduction}

\begin{figure}[ht]
\centering
\includegraphics[width=0.48\textwidth]
{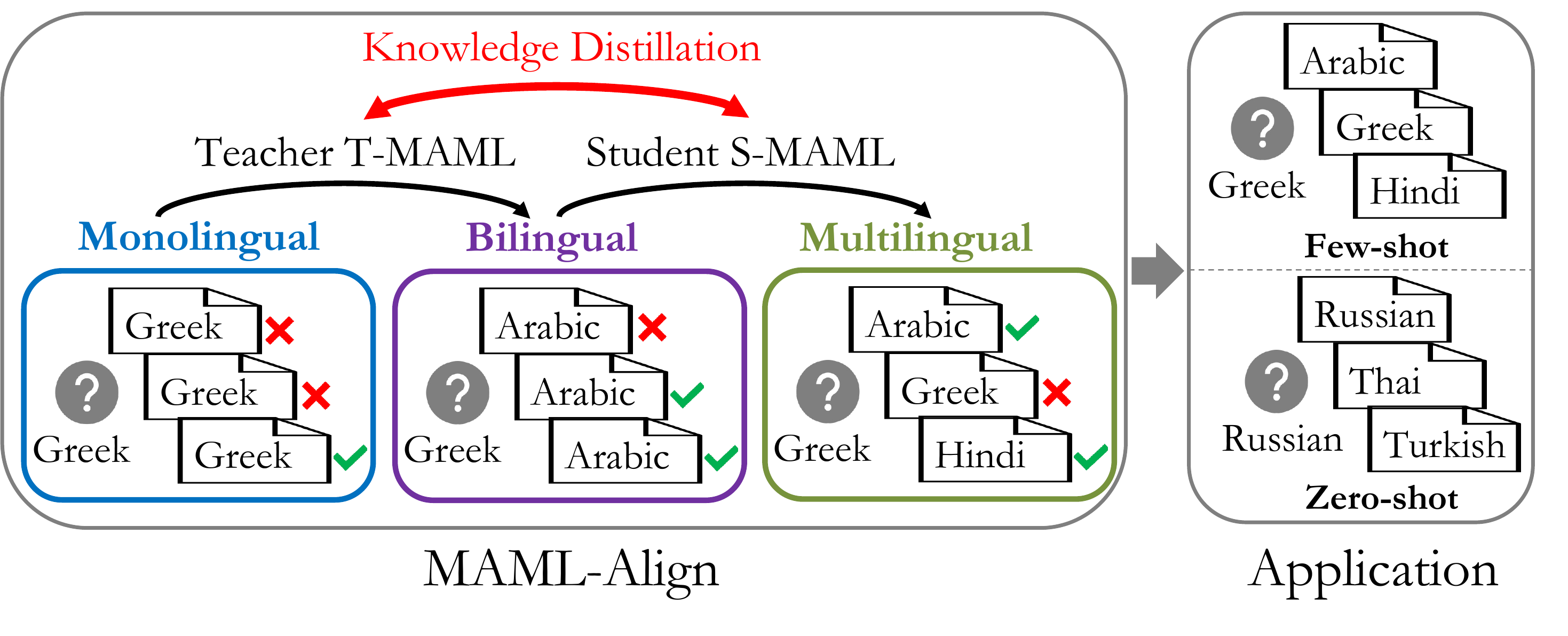} 
\caption{\label{fig:overview-framework} A high-level diagram of our meta-distillation \textbf{\alignmaml{}} framework for multilingual semantic search and some of its application scenarios. We use \lareqa{}~\cite{roy-etal-2020-lareqa} retrieval-based question answering as our benchmark, where the task is to rank and retrieve the most relevant content. We gradually transfer from most to least resourced variants of semantic search. We leverage knowledge distillation to align between the teacher \textbf{\teachmaml{}}~\cite{maml-finn-icml17}, specialized in transferring from monolingual to bilingual, and the student \textbf{\studmaml{}} specialized in transferring from bilingual to multilingual semantic search. The applications can either be few-shot or zero-shot depending on the language arrangements used in the evaluation and whether they are used at any stage in \alignmaml{}. %For that purpose, we curate two support sets and one query set, where the second support set is used as both a query and support set in \teachmaml{} and \studmaml{}, respectively. In the outer loop, we optimize jointly over the distillation and task-specific losses of the query sets.}
}
\vspace{-0.74cm}
\end{figure}

% Paragraph 1: Textual Semantic Search: Context, Pr oblematic, vague introduction with some citations

%% Textual semantic search and how challenging it is or its application or its relation to information retrieval
%% Maybe introduce different types of semantic search: asymmetric and symmetric semantic search
%% Bilingual and multilingual semantic search and how it differs or how challenging it is compared to monolingual semantic search 
%% Focus on one aspect of multilingual semantic search that wasn't tackled well and you wish to tackle better in this paper

Nowadays, the web offers a wealth of information from multiple sources and in different languages. This makes it increasingly challenging to retrieve reliable information efficiently and accurately. Users across the globe may express the need to retrieve relevant content in languages different from the language of the query or in multiple languages simultaneously. All this burgeons the great demand for multilingual semantic search. Compared to bilingual semantic search, often portrayed as cross-lingual information retrieval~\cite{Savoy2019, DBLP:journals/corr/abs-2111-05988}, multilingual or mixed-language semantic search is under-explored and more challenging. It requires not only more semantic understanding but also a stronger alignment between the languages of the query and the contents to be retrieved~\cite{roy-etal-2020-lareqa}.

% This is even more challenging due to the plethora of different languages that exist that the searchers can use while querying. Sometimes, the searcher may use a low-resource language in which there are no relevant documents or may need critical contents expressed in multiple languages at the same time to look at the information from different perspectives. All this burgeons the great demand for multilingual semantic search. Unlike monolingual or bilingual semantic search, multilingual or mixed-language semantic search is under-explored and more challenging as the contents to be retrieved and query can be from any language. % With more than 7,000 spoken languages, existing cross-lingual representations are still not good enough to generalize equally to different downstream tasks. 

% Paragraph 2: Previous work => existing approaches over-reliance on machine translation (look at related work and summarize stuff from there) % Group with the previous paragraph: Previous work on cross-lingual transfer learning approaches 
The new wave of multilingual semantic search focuses on reducing the need to machine translation through transfer learning. Pre-trained multilingual Transformer-based models such as \mbert{}~\cite{bert-devlin-naacl19} and \xlmr{}~\cite{conneau-etal-2020-unsupervised} have been used as off-the-shelf encoders in multilingual semantic search. However, their performance, especially for ad-hoc semantic search, is still lacking~\cite{litschko-multienc-infretr2022}. Knowledge distillation and contrastive-distillation learning approaches are considered as the de-facto approaches to produce better-aligned multilingual sentence representations with reduced need to parallel corpora~\cite{reimers-gurevych-2020-making, tan-multicontrastive-eacl23}. However, they still rely on medium-scaled data including monolingual corpora and back-translation and yield mixed results. Meta-transfer learning, another technique for low-resource learning, has been leveraged for retrieval tasks; however, its application has been restricted to the monolingual case~\cite{lin-chen-2020-preliminary,laadan2019rankml,carvalho2008meta}. Hybrid approaches of meta-learning and knowledge distillation either involve using meta-learning to improve the student-teacher feedback loop ~\cite{zhou-bertlearns2teach-acl22,liu-metakd-arxiv22}, or leverage knowledge distillation to enhance the portability of MAML networks~\cite{zhang-kdf4maml-ecai20}. To the best of our knowledge, we are the first to adapt multilingual meta-transfer learning and to extend an approach based on meta-distillation learning to multilingual semantic search and to a multilingual application in general.  
%  In our work, we investigate other ways to further improve the transfer between different languages on top of semantic specialization in a data-efficient manner. 

% Maybe also group with the previous paragraph: Cross-lingual Meta-transfer learning and its gains 
%% talk about cross-lingual meta-transfer learning in general
%% talk about multilingual meta-transfer learning and how it is under-explored
%% shortcomings to your current approach

% % Group this paragraph with the latter paragraph: Shortcomings of meta-transfer learning and knowledge distillation and how combining them could help 

% Contributions:
%% Present the approach with more overview details => present Figure 1 too
%% Describe to which benchmarks this is applied
%% Present some key findings 
In this paper, inspired by \citet{mhamdi-etal-2021-x}, which propose the X-METRA-ADA algorithm to adapt meta-learning to cross-lingual transfer learning for cross-lingual natural language understanding, we propose an adaptation of meta-transfer learning to multilingual semantic search. Given the lack of resourcefulness of semantic search especially in the multilingual case, this encourages us to pursue a meta-learning direction based on MAML. We also explore the combination of meta-learning and knowledge distillation and adapt it to the task of multilingual semantic search (Figure~\ref{fig:overview-framework}). We do that in two stages 1) from monolingual to bilingual and 2) from bilingual to multilingual to create a more gradual feedback loop, which makes it easier to generalize to the multilingual case. We conduct experiments on different semantic search benchmarks on top of a strong baseline based on sentence transformers~\cite{reimers-gurevych-2019-sentence}. Our findings confirm the benefits of the meta-distillation approach compared to naive fine-tuning and \maml{}.

Our \textbf{main contributions} are:
\begin{enumerate*}[label=(\arabic*)]
    \item {We are the first to propose a meta-learning approach for multilingual semantic search~(\S\ref{subsec:upstream-models}) and to curate meta-tasks for that effect~(\S\ref{subsec:meta-learning-stages}).}
    \item {We are the first to propose a meta-distillation approach to distill the transfer from monolingual to bilingual to the transfer from bilingual to multilingual semantic search~(\S\ref{para:alignmaml}).}
    \item{We systematically compare between several few-shot transfer learning methods and show the gains of our multilingual meta-distillation approach~(\S\ref{subsec:multi-bil-mono-perf}).}
    %\item {We systematically compare meta-learning on two semantic search benchmarks to different baselines including machine translation(~\S\ref{sec:experimental-setup}).}
    \item{We also conduct ablation studies involving different language arrangements and different sampling approaches in the meta-task construction~(\S\ref{sec:ablation-studies}).}
\end{enumerate*}
\vspace{-0.3cm}

% low-resource multilingual semantic search
% meta-learning helps 
% alignment works query and support sets are together 

\section{Related Work}
% Summary of machine translation and cross-lingual transfer learning for semantic search and information retrieval \cite{DBLP:journals/corr/abs-2111-05988}
\vspace{-0.2cm}

\paragraph{Transfer Learning for Multilingual Semantic Search}

Most approaches to multilingual semantic search or cross-lingual information retrieval rely on machine translation to reduce the problem to monolingual search~\cite{lu-querytrans-aclclp08,nguyen-wikiTranslate-clef08,jones-domainspecificquery-ijcnlp08}. However, such systems are inefficient for multilingual semantic search due to error propagation and overheads from API calls. In addition to that, the number of language combinations in the query and content to be retrieved can get prohibitively large~\cite{Savoy2019}. More prominent approaches leverage transfer learning with models like \mbert{} and \xlm{} used for question-answer retrieval~\cite{76-DBLP:conf/acl/YangCAGLCAYTSSK20}, bitext mining~\cite{85-DBLP:conf/lrec/ZiemskiJP16, 86-DBLP:conf/lrec/ZweigenbaumSR18a}, and semantic textual similarity~\cite{26-DBLP:conf/adcs/HoogeveenVB15, 35-DBLP:conf/naacl/LeiJBJTMM16} and show that semantic specialization and pre-fine-tuning on other auxiliary tasks helps. 

\vspace{-0.2cm}
\paragraph{Multilingual Meta-Learning}

Meta-transfer learning, or ``learning to learn'' has found favor in cross-lingual transfer learning for numerous downstream applications~\cite{DBLP:conf/emnlp/GuWCLC18,DBLP:conf/icassp/HsuCL20,DBLP:conf/acl/WinataCLLXF20, DBLP:conf/interspeech/ChenHLL20, DBLP:conf/aaai/XiaoGZZLL21}. Most recent meta-learning work involving transferring between different languages focuses on cross-lingual meta-learning~\cite{nooralahzadeh-etal-2020-zero,mhamdi-etal-2021-x}. Meta-transfer learning has been extended multilingually by exploring joint multi-task and multi-lingual transfer~\cite{tarunesh-metamulti-acl21,van2021multilingual}. %~\citet{tarunesh-metamulti-acl21} propose a meta-learning framework for multi-task and multi-lingual transfer and show that this joint approach enables effective sharing of parameters across multiple tasks and multiple languages.~\citet{van2021multilingual} propose a meta-learning framework and show its effectiveness in both the cross-lingual and multilingual training adaptation settings of document classification. However, their multilingual evaluation is focused on the scenario where the same target languages during meta-testing can be also used as auxiliary languages during meta-training. This motivates us to explore multilingual meta-transfer learning for semantic search and to test the generalizability of our meta-learning model even for unseen languages.

\vspace{-0.3cm}

\paragraph{Meta-Distillation Learning}
Meta-learning has also been leveraged to improve the performance of knowledge distillation to help the teacher transfer better to the student~\cite{zhou-bertlearns2teach-acl22, liu-metakd-arxiv22}. Inversely, knowledge distillation has been leveraged to improve meta-learning, especially MAML, by making it more portable~\cite{zhang-kdf4maml-ecai20}. \citet{xu-gradualtuning-aclwork2021} follow a gradual multi-stage process which is different in scope and approach from our work in that it uses fine-tuning for domain adaptation to interpolate between in-domain and out-domain data. In contrast, we apply our approach to a multilingual semantic search in an end-to-end meta-learning framework which gradually meta-transfers between semantic search language variants. Moreover, we show that our approach outperforms naive joint fine-tuning, advocating for a meta-learning approach in the few-shot learning scenario.\footnote{More detailed related work can be found in Appendix~\ref{app:more-related-work}.}

\vspace{-0.3cm}
%% how your paper combines all of that

%%% cite cross-task cross-lingual transfer
%%% Cross-Lingual Transfer with MAML on Trees
%%% Soft Layer Selection with Meta-Learning for Zero-Shot Cross-Lingual Transfer
%%% Meta-Learning for Fast Cross-Lingual Adaptation in Dependency Parsing
%%% Meta-XNLG: A Meta-Learning Approach Based on Language Clustering for Zero-Shot Cross-Lingual Transfer and Generation
%%% Meta-Learning a Cross-lingual Manifold for Semantic Parsing
%%% Meta-ED: Cross-lingual Event Detection using Meta-learning for Indian Languages

%%% Meta-Learning for Effective Multi-task and Multilingual Modelling
%%% Multilingual and cross-lingual document classification: a meta-learning approach
%%% "Diversity and Uncertainty in Moderation" are the Key to Data Selection for Multilingual Few-shot Transfer

%% Describe ACL papers in the area
%%% MetaTS: Meta Teacher-Student Network for Multilingual Sequence Labeling with Minimal Supervision
%%% ACL 2022 papers on meta-distil in general

\section{Meta-Learning Background}
\label{sec:background}

% \subsection{Meta-Learning}
% \label{sec:back-metalearning}
%In the dawn of conventional machine learning, learning is more focused on optimizing on examples from a given training dataset for a particular task. 
Given a training dataset $\mathcal{D}$ made of instances: $\{(x_{1}, y_{1}),\dotsc, (x_{n}, y_{n})\}$, the goal of a conventional machine learning model is to find the most optimal parameters $\theta^*$ that minimize the loss $\mathcal{L}$: 
\begin{equation}
    \theta^{*} = \underset{\theta}{\arg\min} \mathcal{L}(\theta;\omega;\mathcal{D}),
\label{eq:conv-ml}
\end{equation}
where $\omega$ is some already acquired prior knowledge or assumption on how to learn \cite{meta-survey-20}. There are two main distinctions between this conventional machine-learning process and meta-learning. First, machine learning focuses on one task at a time whereas meta-learning optimizes over a distribution of many sub-tasks, referred to as 'meta-tasks', sampled to simulate a low-resource scenario.
Second, meta-learning effectively learns the prior knowledge jointly with the task by adding an extra layer of abstraction to the process.

%From a \ul{task-distribution perspective}, meta-learning is a mechanism to learn a general-purpose algorithm of how to generalize across different meta-tasks. Unlike conventional tasks, meta-tasks are made up of many simulations of the same task. Thus, meta-learning enables learning under low-resource scenarios. 

Each meta-task is defined as a tuple $T = (S, Q)$, where $S$ and $Q$ denote support and query sets, respectively. $S$ and $Q$ are sampled to simulate the train and test labeled subsets of instances. Following a bi-level optimization abstraction (as in MAML), the meta-learning process is a sequence of inner loops each followed by an outer loop. The inner loop is specialized in learning task-specific optimizations over the support sets in a batch of meta-tasks; the outer loop, on the other hand, learns the generalization over the query sets in the same batch in a leader-follower manner. The goal is to learn a proper initialization point to generalize to the domain of $Q$. Meta-learning works with meta-training $\metatraindata = \{ \metatrainspt, \metatrainqry\}$, meta-testing $\metatestdata = \{ \metatestspt, \metatestqry\}$, and optionally meta-validation $\metavaliddata$ datasets. During \textbf{meta-training}, we start by learning the optimal prior knowledge $\omega^{*}$:
\begin{equation}
    \omega^{*} = \underset{\omega}{\arg\min} \mathcal{L}(\omega|\metatraindata).
\end{equation} 
This learned prior knowledge is leveraged along with  the support set in the meta-testing dataset $\metatestspt$ during \textbf{meta-testing} to fastly adapt to $\metatestqry$ (without optimizing on it like in meta-training), as follows:
\begin{equation}
    \theta^{*} = \underset{\theta}{\arg\min} \mathcal{L}(\theta|\omega^{*}, \metatestspt).
    \label{eq:meta-learn}
\end{equation}

% \vspace{-0.2cm}
% \begin{itemize}[leftmargin=*]
%     \itemsep0em
%     \item{\textbf{Meta-training dataset ($\metatraindata$)}. This includes meta-tasks sampled using the $\train{}$ split. During meta-training, our models optimize over both the support and the query set of each meta-task in the bi-level optimization process.}
%     \item{\textbf{Meta-validation dataset ($\metavaliddata$)}. This is similar to meta-training except that the $\dev{}$ split is used in this case.}
%     \item{\textbf{Meta-testing dataset ($\metatestdata$)}. The $\test{}$ split is used here. The meta-trained and validated model undergoes additional passes over the support set to compute the performance on the query without updating the learner parameters.}
% \end{itemize}

% Compared with Equation \ref{eq:conv-ml}, Equation \ref{eq:meta-learn} is conditioned on the prior knowledge (i.e. meta-knowledge) which is already learned. 
\vspace{-0.3cm}

%It is also worth noting the distinction between using all acquired knowledge coming from all seen support tasks to apply it directly to a new set versus using the acquired meta-knowledge to know how to learn from a new support set $\metatestspt$. The former case is undesirable and is denoted as over-fitting and shifts away from the promised task generalization that we are strongly seeking. 

\section{Methodology}
\label{sec:methodology}

In this section, we start by defining the task of sentence-level semantic search and its different categories~(\S\ref{subsec:task-formulation}), its language variants~(\S\ref{sec:task-lang-variants}), and supervision degrees~(\S\ref{sec:supervision-degrees}). Then, we present our optimization-based meta-distillation learning algorithm \alignmaml{} and show how it extends from the original \maml{} algorithm~(\S\ref{subsec:upstream-models}).
\vspace{-0.1cm}
\subsection{Task Formulation}
\label{subsec:task-formulation}

Our base task is sentence-level semantic search. Given a sentence query $q$ from a pool of queries $\allqueries$, the goal is to find relevant content $r$ from a pool of candidate contents $\allcontent$. The queries are of sentence length and retrieved contents are either sentences or small passages of few sentences. 

% Our base task is sentence-level semantic search. Given a sentence query $q$ in a language $\lng \in \alllang$, the goal is to find relevant content $r$ from a set of contents $\allcontent$ in one or more languages. The queries are of sentence length and retrieved contents can include sentences or small passages of few sentences and can belong to multiple languages for a particular query in each language.

In terms of the format of the queries and contents, there are two main categories of semantic search: 
%\begin{itemize}[leftmargin=*]
\begin{enumerate*}[label=(\arabic*)]
\itemsep0em
\vspace{-0.25cm}
    \item{\textbf{Symmetric Semantic Search.} Query $q$ and relevant content $r$ have approximately the same length and format.}
    \item{\textbf{Asymmetric Semantic Search.} $q$ and $r$ are not of the same length or format. For example, $q$ can be a question and $r$ a passage answering that.}
%\end{itemize}
\end{enumerate*}

\vspace{-0.3cm}

\subsection{Task Language Variants}
\label{sec:task-lang-variants}

In the context of languages, we distinguish between three variants of semantic search at evaluation time (also illustrated in Figure~\ref{fig:overview-framework}):   
% \vspace{-0.2cm}
% \begin{itemize}[leftmargin=*]
\begin{enumerate*}[label=(\arabic*)]
\itemsep0em
\item{\textbf{Monolingual Semantic Search (\mono{}).} The pools of queries and candidate contents $\allqueries$ and $\allcontent$ are from the same known and fixed language $\lng_{\allqueries} = \lng_{\allcontent} \in \alllang$.}
% \vspace{-0.1cm}
\item{\textbf{Bilingual Semantic Search (\bil{}).}   The pools of queries and candidate contents are sampled from two different languages $\{\lng_{\allqueries}, \lng_{\allcontent}\} \in \alllang^{2}$, such that $\lng_{\allqueries} \neq \lng_{\allcontent}$.} % This is also termed as cross-lingual semantic search~\cite{Savoy2019}.
% \vspace{-0.1cm}
\item{\textbf{Multilingual Semantic Search (\multi{}).} This is the problem of retrieving relevant contents from a pool of candidates from a subset of multiple languages $\alllang_{\allcontent} \subseteq \alllang$ to a query expressed in a subset of multiple languages $\alllang_{\allqueries} \subseteq \alllang$.} % queries $\lng_q \in \alllang$ candidate contents $\lng_r \in \alllang$
% \end{itemize}
\end{enumerate*}
% \vspace{-0.2cm}

Unlike monolingual and bilingual semantic search, multilingual semantic search doesn't restrict or condition on which languages can be used in the queries or the candidate contents. Therefore, it is more challenging and requires stronger multilingual alignment~\cite{roy-etal-2020-lareqa}. 

\subsection{Supervision Degrees}
\label{sec:supervision-degrees}
In the absence of enough training data for the task, we distinguish between three degrees of supervision of semantic search:   
\vspace{-0.2cm}
\begin{itemize}[leftmargin=*]
\itemsep0em
\item{\textbf{Zero-Shot Learning.} This resembles ad-hoc semantic search in that it doesn't involve any fine-tuning specific to the task of semantic search. Rather, off-the-shelf pre-trained language models are used directly to find relevant content to a specific query. This still uses some supervision in the form of parallel sentences used to pre-train those off-the-shelf models. In the context of multilingual semantic search, we include in the zero-shot learning case any evaluation on languages not seen during fine-tuning.}
\vspace{-0.1cm}
\item{\textbf{Few-Shot Learning.} Few-shot learning is used in the form of a small fine-tuning dataset. In the context of multilingual semantic search, we talk about a few-shot evaluation for any language seen either in the arrangement of the query or the contents to be retrieved during fine-tuning.}
\end{itemize}

\vspace{-0.3cm}

\subsection{Meta-Learning Models}
\label{subsec:upstream-models}

\paragraph{Original \maml{} Algorithm.}
\label{para:maml}

Our first variant is a direct adaptation of MAML to multilingual semantic search. We use the procedure outlined in Algorithm~\ref{algo-tmaml}. We start by sampling a batch of meta-tasks from a meta-dataset distribution $\setset{\mbox{\tiny{X$\shortrightarrow$ $\xprime{X}$ }}}$, which simulates the transfer from $X$ to $\xprime{X}$. $X$ and $\xprime{X}$ denote different task language variants of semantic search (monolingual, bilingual, multilingual, or any combination of that). We start by initializing our meta-learner parameters $\theta$ with the pre-trained learner parameters $\theta_B$. For each meta-batch, we perform an inner loop (Algorithm~\ref{procedure-inner-loop}) over each meta-task $T_j=(S_j, Q_j)$, separately, where we update $\theta_{j}$ using $S_j^X$ for $n$ steps. At the end of the inner loop, we compute the gradients with respect to the loss of $\theta_j$ on $Q_j^{\xprime{X}}$. After finishing a pass over all meta-tasks of the batch, we perform one outer loop by summing over all pre-computed gradients and updating $\theta$. 

\begin{algorithm}[ht]
\caption{\maml{}: Transfer Learning from X to $\xprime{X}$ (X$\rightarrow$ $\xprime{X}$)% : Model-Agnostic Meta-Learning for Cross-lingual Transfer
}
\begin{small}
\begin{algorithmic}[1]
% \setxtoy{X}{\xprime{X}}
% \newcommand{\setxtoy}[2]{\setset{\mbox{\tiny{#1 \rightarrow #2}}}}
\Require Task set distribution $\setset{\mbox{\tiny{X $\shortrightarrow \xprime{X}$}}}$ simulating transfer from X to $\xprime{X}$ task language variants, pre-trained learner $B$ with parameters $\theta_{B}$, and meta-learner $M$ with parameters ($\theta$, $\alpha$, $\beta$, $n$).
\State Initialize $\theta \leftarrow \theta_{B}$ 
\While{not done}
\State Sample a batch of tasks $\mathcal{T}= \{T_1, \ldots T_b\} \sim \setset{\mbox{\tiny{X$\shortrightarrow \xprime{X}$}}}$

\State $\mathcal{L}^{S_j^X}_{T_j}, \mathcal{L}^{Q_j^{\xprime{X}}}_{T_j}(B_{\theta_{j}}) $= INNER\_LOOP($\mathcal{T}$, $\theta$, $\alpha$, $n$)

\State Outer Loop: Update $\theta \leftarrow \theta - \beta \nabla_{\theta} \sum^{b}_{j=1} \mathcal{L}^{Q_j^{\xprime{X}}}_{T_j}(B_{\theta_{j}}) $ 
\EndWhile

\end{algorithmic}
\end{small}
\label{algo-tmaml}
\end{algorithm}
\begin{algorithm}[ht]
\caption{INNER\_LOOP
}
\begin{small}
\begin{algorithmic}[1]

% \textbf{Function} {INNER\_LOOP}($\mathcal{T}$, $\theta$, $\alpha$, $n$)%\textbf{Function} {INNER\_LOOP}($\mathcal{T}$, $\theta$, $\alpha$, $n$)

\Function{INNER\_LOOP}{$\mathcal{T}$, $\theta$, $\alpha$, $n$}
   \For{each $T_j = (S_j^{X}, Q_j^{\xprime{X}})$ in $\mathcal{T}$}
    \State Initialize $\theta_{j} \leftarrow \theta$ 
    \For{$t=1\ldots n$}
    \State Evaluate $\partial B_{\theta_{j}} / \partial \theta_{j}  = \nabla_{\theta_{j}} \mathcal{L}^{S_j^X}_{T_{j}}(B_{\theta_{j}})$
    \State Update $\theta_{j} = \theta_{j} - \alpha \partial B_{\theta_{j}} / \partial \theta_{j} $
    
    \EndFor
    
    \State Evaluate query loss $\mathcal{L}^{Q_j^{\xprime{X}}}_{T_j}(B_{\theta_{j}}) $ and save it for outer loop
    \EndFor
\EndFunction
\end{algorithmic}
\end{small}
\label{procedure-inner-loop}
\end{algorithm}

% \vspace{-0.5cm}

\paragraph{\alignmaml{} Algorithm.}
\label{para:alignmaml}

The idea behind this extension is to use knowledge distillation to distill \teachmaml{} to \studmaml{} and improve the generalization of \maml{} in Algorithm~\ref{algo-maml2maml} across different modes of transfer.  \teachmaml{} is more high-resource  than \studmaml{}. Given meta-tasks from $\setset{\mbox{\tiny{X$\shortrightarrow$Y}}}$ and $\setset{\mbox{\tiny{Y$\shortrightarrow$Z}}}$, the goal is to use that shared mode of transfer $Y$ to align different modes of transfer of semantic search. After executing the two inner loops of the two \maml{}s (with more inner steps for \teachmaml{} than \studmaml{}), where the support sets are sampled from $X$ and $Y$, respectively, we compute in the optimization process of the outer loop: the weighted combination of \textbf{$\mathcal{L}_{task}$}, the average over the task-specific losses on the query sets sampled from $Y$ and $Z$, and \textbf{$\mathcal{L}_{kd}$}, the distillation loss on $Y$.

\begin{algorithm}[ht]
\caption{\alignmaml{}: Knowledge distillation to align two different \maml{}s (X$\rightarrow$Y$\rightarrow$Z)
}
\begin{small}
\begin{algorithmic}[1]
\Require Task set distribution $\setset{\mbox{\tiny{X$\shortrightarrow$Y}}}$ and $\setset{\mbox{\tiny{Y$\shortrightarrow$Z}}}$ sharing the same $Y$,  pre-trained learner $B$ with parameters $\theta_{B}$, and meta-learners $M{\mbox{\tiny{X $\shortrightarrow$ Y}}}$ with parameters ($\theta$, $\alpha$, $\beta$, $n$) and $M{\mbox{\tiny{Y $\shortrightarrow$ Z}}}$ with parameters ($\xprime{\theta}$, $\alpha$, $\xprime{\beta}$, $\xprime{n}$), where $\xprime{n} < n$.
\State Initialize $\theta \leftarrow \theta_{B}$ 
\State Initialize $\xprime{\theta} \leftarrow \theta_{B}$
\While{not done}
% Do it batch by batch
\State Sample batch of tasks $\mathcal{T_{\mbox{\tiny{X$\shortrightarrow$Y}}}}=\{T_1, \ldots T_b\} \sim \setset{\mbox{\tiny{X$\shortrightarrow$Y}}}$  

\State Sample batch of tasks $\mathcal{T_{\mbox{\tiny{Y$\shortrightarrow$Z}}}}= \{T_1, \ldots T_b\} \sim \setset{\mbox{\tiny{Y$\shortrightarrow$Z}}}$

%\STATE Sample batch of tasks $\mathcal{T_{\mbox{\tiny{X$\shortrightarrow$Z}}}}= \{T_1, \ldots T_b\} \sim \setset{\mbox{\tiny{X$\shortrightarrow$Z}}}$ 

\State
$\mathcal{L}^{S_j^X}_{T_j}, \mathcal{L}^{Q_j^Y}_{T_j}$ = INNER\_LOOP($\mathcal{T_{\mbox{\tiny{X$\shortrightarrow$Y}}}}$, $\theta$, $\alpha$, $n$)

%\STATE $\mathcal{L}^{Q_j^{\xprime{Z}}}_{T_j}(B_{\theta_{j}})$ = INNER\_LOOP($\mathcal{T_{\mbox{\tiny{X$\shortrightarrow$Y}}}}$, $\xprime{\theta}$, $\xprime{\alpha}$)

\State $\mathcal{L}^{S_j^Y}_{T_j},\mathcal{L}^{Q_j^Z}_{T_j}$ = INNER\_LOOP ($\mathcal{T_{\mbox{\tiny{Y$\shortrightarrow$Z}}}}$, $\xprime{\theta}$, $\alpha$, $\xprime{n}$)

% \FOR{all $T_j = (S_j^X, Q_j^Y)$ in $\mathcal{T}$}
% \STATE Initialize $\theta_{j} \leftarrow \theta$ 
% \FOR{$t=1\ldots n$}
% \STATE Evaluate $\partial B_{\theta_{j}} / \partial \theta_{j}  = \nabla_{\theta_{j}} \mathcal{L}^{S_j^X}_{T_{j}}(B_{\theta_{j}})$
% \STATE Update $\theta_{j} = \theta_{j} - \alpha \partial B_{\theta_{j}} / \partial \theta_{j} $
% \ENDFOR
% \STATE Evaluate query loss $\mathcal{L}^{Q_j^Y}_{T_j}(B_{\theta_{j}}) $ and save it for outer loop
% \ENDFOR

% \FOR{all $T_j = (S_j^X, Q_j^Y)$ in $\mathcal{T}$}
% \STATE Initialize $\theta_{j} \leftarrow \theta$ 
% \FOR{$t=1\ldots n$}
% \STATE Evaluate $\partial B_{\theta_{j}} / \partial \theta_{j}  = \nabla_{\theta_{j}} \mathcal{L}^{S_j^X}_{T_{j}}(B_{\theta_{j}})$
% \STATE Update $\theta_{j} = \theta_{j} - \alpha \partial B_{\theta_{j}} / \partial \theta_{j} $
% \ENDFOR
% \STATE Evaluate query loss $\mathcal{L}^{Q_j^Y}_{T_j}(B_{\theta_{j}}) $ and save it for outer loop
% \ENDFOR

% with respect to the weighted combination of the task-specific losses and the distillation loss
%\STATE Update $\theta \leftarrow \theta - \beta [\nabla_{\theta} \sum^{b}_{j=1} \mathcal{L}^{Q_{j}}_{T_j}(B_{\theta_{j}}) - \lambda  \nabla_{\theta} \sum^{b}_{j=1} KL(\mathcal{L}^{Q_{j}^Z}_{T_j}(B_{\theta_{j}}), \mathcal{L}^{Q_j^{\xprime{Z}}}_{T_j}(B_{\theta_{j}}))] $

\State $\mathcal{L}_{task} = \sum^{b}_{j=1} \frac{\mathcal{L}^{Q^Y_j}_{T_j}(B_{\theta_{j}}) + \mathcal{L}^{Q^Z_j}_{T_j}(B_{\theta_{j}})}{2}$

\State $\mathcal{L}_{kd} = KL(\sum^{b}_{j=1}\mathcal{L}^{Q_j^Y}_{T_j}(B_{\theta_{j}}), \sum^{b}_{j=1}\mathcal{L}^{S_j^Y}_{T_j}(B_{\theta_{j}}))$

\State Update $\theta \leftarrow \theta - \beta \nabla_{\theta}  (\mathcal{L}_{task} + \lambda \mathcal{L}_{kd}) $

\EndWhile
\end{algorithmic}
\end{small}
\label{algo-maml2maml}
\end{algorithm}

\begin{figure}[ht]
\centering
\includegraphics[width=0.5\textwidth]
{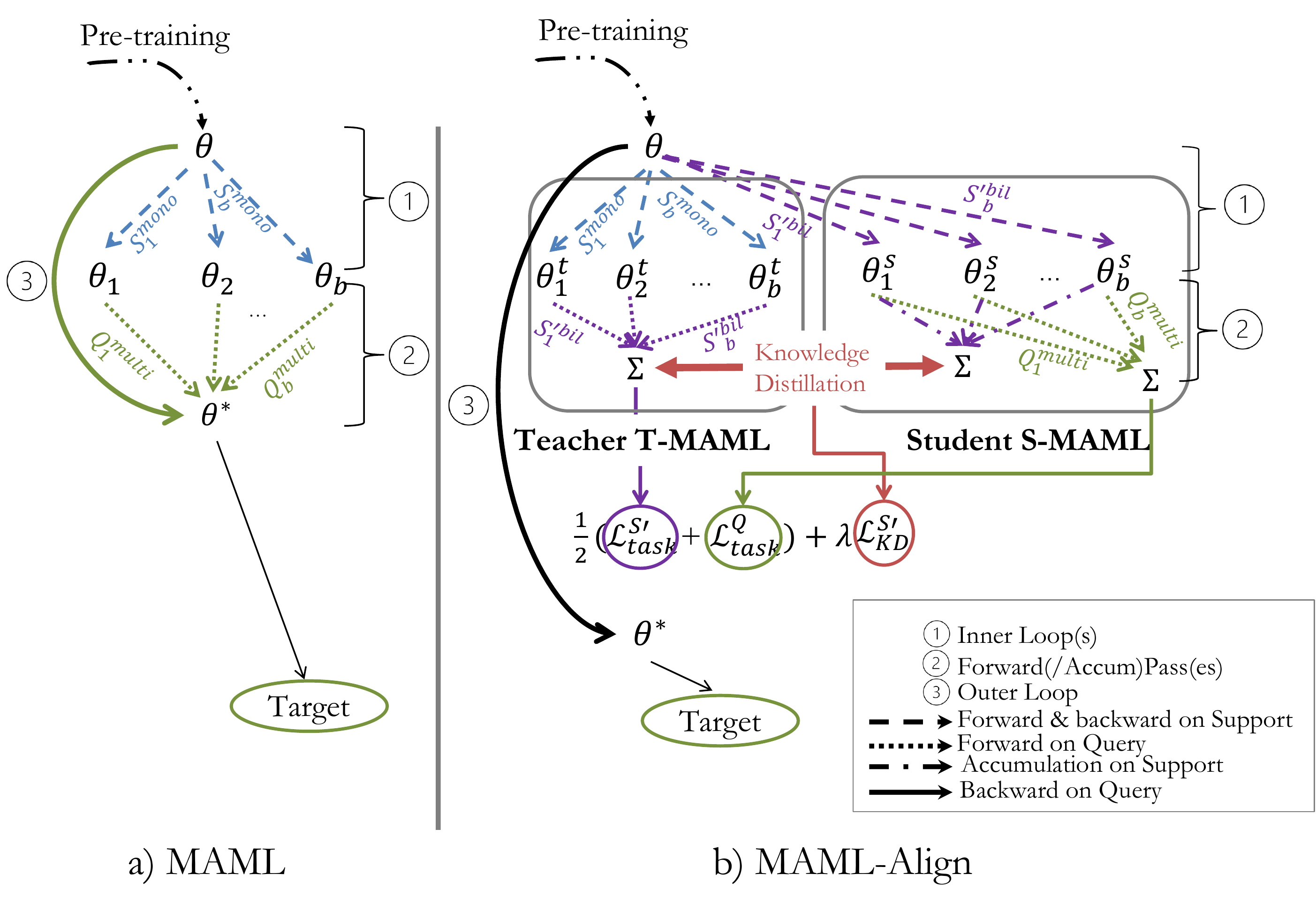}
\caption{\label{fig:model_variants} A conceptual comparison between \textbf{\alignmaml{}} and the original meta-learning baseline \textbf{\maml{}}. A single iteration of \maml{} involves one inner loop optimizing over a batch of support sets from a source language variant of the task followed up by an outer loop optimizing over the batch query sets curated from the target task variant. In \alignmaml{}, on the other hand, we curate two support sets and one query set, where the second support set is used as both a query and support set in \teachmaml{} and \studmaml{}, respectively. We perform two inner loops and two forward passes. Then, in the outer loop, we optimize jointly over the distillation and task-specific losses of the query sets.}
\vspace{-0.65cm}
\end{figure}

Figure~\ref{fig:model_variants} shows a conceptual comparison between \alignmaml{} and \maml{}.

\vspace{-0.1cm}
\section{Experimental Setup}
\label{sec:experimental-setup}
In this section, we describe the downstream datasets and models~(\S\ref{subsec:datasets}), their formulation as meta-tasks~(\S\ref{subsec:meta-learning-stages}), and the different baselines and model variants used in the evaluation~(\S\ref{subsec:baselines}). 
\vspace{-0.2cm}
\subsection{Downstream Benchmarks}
\label{subsec:datasets}

% from XTREME-R~\cite{ruder-etal-2021-xtreme}, and semantic similarity
We evaluate our proposed approaches over the following combination of multilingual and bilingual sentence-level semantic search datasets for which we describe the downstream models used:\footnote{More details on the base model architectures can be found in Appendix~\ref{app:base-models} More experimental details on the datasets are and hyperparameters used in Appendix~\ref{app:more-experimental-setup}.}
\vspace{-0.3cm}
\begin{itemize} [leftmargin=*]
\itemsep0em
\item{\textbf{Asymmetric Semantic Search.} We use LAReQA~\cite{roy-etal-2020-lareqa}, focusing on XQuAD-R, which is a retrieval-based task reformulated from the span-based question answering XQuAD~\cite{artetxe-etal-2020-cross}. This dataset covers 11 languages. In this work, we only use seven languages: Arabic, German, Greek, Hindi (used for few-shot learning), Russian, Thai, and Turkish (kept for zero-shot evaluation).\footnote{We download the data from \url{https://github.com/google-research-datasets/lareqa}.} We design a Transformer-based triplet-encoder model(modified from the original dual encoder in~\citet{roy-etal-2020-lareqa}) with three towers encoding 1) the question, 2) its answer and its context, and 3) the negative candidates and their contexts.} % We don't use MLQA-R~\cite{lewis-etal-2020-mlqa}, which is not fully parallel like XQuAD-R.
\vspace{-0.1cm}
\item{\textbf{Symmetric Semantic Search.} As there is no multilingual parallel benchmark for symmetric search, we focus, in our few-shot learning experiments, on a small-scale bilingual benchmark. We use \stsbm{} from SemEval-2017 Task 1~\cite{cer-etal-2017-semeval}.\footnote{We download SemEval-2017 evaluation and its ground truth scores from \url{https://alt.qcri.org/semeval2017/task1/index.php?id=data-and-tools}, which covers English-English, Arabic-Arabic, Spanish-Spanish, Arabic-English, Spanish-English, and Turkish-English.} This is a semantic similarity benchmark, which consists of a collection of sentence pairs drawn mostly from news headlines. Each sentence pair is scored between 1 and 5 to denote the extent of their similarity. We use a Transformer-based dual-encoder model, which encodes sentences 1 and 2 in each sentence pair using the same shared encoder and computes the cosine similarity score.}
\end{itemize}
\vspace{-0.5cm}

% \begin{table}[h]
% \centering
% \scalebox{0.95}{
% \small
% \begin{tabular}{l|l|ll|ll|ll}
% \toprule
% \textbf{Lang} & \textbf{ISO} & \textbf{\# Sentences} \\
% \toprule
% Telugu  & TE & 234 \\
% Javanese & JV & 205 \\
% Kazakh & KK & 575 \\
% Thai & TH & 548 \\
% Malayalam & ML & 687 \\
% Tamil & TA & 307 \\
% \bottomrule
% \end{tabular}
% }
% \caption{Statistics of TATEOBA test split whole for languages which don't have 1,000 aligned sentences.}
% \label{tab:tatoeba-stats}
% \end{table}

\subsection{Meta-Datasets}
\label{subsec:meta-learning-stages}
Following our formulation of semantic search downstream benchmarks, we construct pseudo-meta-tasks by drawing from the available triplets or sentence pairs to form the support set $S$, so that each support set consists of a batch of $k\_shot$ triplets or sentence pairs. Then, we form the triplets or sentence pairs in the query set $Q$ by picking for each question or sentence pair in $S$ either a similar or random question or sentence pair. Details of the different transfer modes and their support and query set arrangements are in Table~\ref{tab:meta-tasks} in Appendix~\ref{app:meta-tasks}. We construct meta-datasets for different stages of meta-learning where \train{}, \dev{}, and \test{} splits are used to sample $\metatraindata$, $\metavaliddata$, and $\metatestdata$, respectively. The optimizations on $\metatraindata$ and $\metatestdata$ are as defined in~\S\ref{sec:background} and the optimization on $\metavaliddata$ is similar to that of $\metatraindata$.

\vspace{-0.3cm}

\subsection{Baselines \& Model Variants}
\label{subsec:baselines}

%In order to systematically and consistently evaluate our meta-learning approaches, we design a series of experiments. 
Since we are the first, to the best of our knowledge, to explore meta-learning for bilingual or multilingual information retrieval or semantic search, we only compare with respect to our internal variants and include external non-meta-learning baselines. 
\vspace{-0.3cm}
\paragraph{Baselines.}
\label{para:baselines}
We design the following baselines:
\vspace{-0.3cm}
\begin{itemize}[leftmargin=*]
    \itemsep0em
    \item{\textit{BASE}: This is our initial zero-shot approach based on an off-the-shelf pre-trained language model. For the rest of our analysis, we use the best model on our 5-fold cross-validation test splits, which is sentence-BERT (\sbert{}) paraphrase-multilingual-mpnet-base-v2, according to our preliminary evaluation of different Sentence Transformers models.\footnote{\url{https://huggingface.co/sentence-transformers} in Table~\ref{tab:baseline-models-lareqa} in Appendix~\ref{app:more-experimental-setup}.}} 
    %either either pre-fine-tuned to some high-resource (namely \mbert{}+\ensquad{)} data or with respect to some large-scale optimization based on some alignment loss (namely \sbert{})
    \vspace{-0.2cm}
    \item{\textit{\sbert{}+\finetune{}}: On top of S-BERT, we fine-tune jointly and directly on the support and query sets of each meta-task in $\metatraindata$ and $\metavaliddata$. This few-shot baseline makes for a fair comparison with the meta-learning approaches.}
\end{itemize}
\vspace{-0.4cm}
\paragraph{Internal Variants.}
  \label{para:internal-variants}
   We design the following meta-learning variants:
\vspace{-0.2cm}
\begin{itemize}[leftmargin=*]
    \itemsep0em
\item{\textit{\sbert{}+\maml{}}: On top of \sbert{}, we apply \maml{} (following Algorithm~\ref{algo-tmaml}). At each episode, we conduct a meta-training followed by a meta-validation phase.}
    \vspace{-0.1cm}
\item{\textit{\sbert{}+\alignmaml{}}: On top of \sbert{}, we apply \alignmaml{}{} (following Algorithm~\ref{algo-maml2maml}). Similarly, at each episode, we conduct a meta-training followed by a meta-validation phase.}
\end{itemize}
  \vspace{-0.3cm}
\paragraph{External Evaluation.}
\label{para:external-evaluation}

% \begin{itemize}
 % \item{Machine translation:}
 To assess the impact of using machine translation models with or without meta-learning and the impact of machine translation from higher-resourced data, we explore Translate-Train (\translatetrain{}), where we translate English data in \ensquad{}\footnote{We use the translate.pseudo-test provided for XQuAD dataset by XTREME benchmark \url{https://console.cloud.google.com/storage/browser/xtreme_translations}.} and \stsben{}\footnote{We use the translated dataset from the original English STSB \url{https://github.com/PhilipMay/stsb-multi-mt/}.} to the evaluation languages. We then either use translated data in all languages or in each language separately as a data augmentation technique.
%     \begin{itemize}[leftmargin=*]
% \itemsep0em
%         \item{Translate-Train (\translatetrain{}): We translate the  English high-resource data in \ensquad{} and \stsben{} to evaluation languages. This data is used to train some internal variants to evaluate the impact of machine translation from higher-resourced data.}
        %\item{Translate-Evaluate (\translatetest{}): We use the translation API to translate the evaluation data in \lareqa{} and \stsbm{} in any language back to English. Then, we apply any model in internal variant to the translated data. - for \sbert{}+\translatetest{} means it is not reasonable to report such results since everything is translated to English so we cannot distinguish between multilingual, bilingual, or monolingual evaluations, thus a single number is reported.}
    % \end{itemize}
    %}
    % \item{Knowledge distillation: To evaluate}
% \end{itemize}
\vspace{-0.3cm}

\begin{table*}[t] 
\centering
\scalebox{0.59}{ 
\begin{tabular}{l|l|lll|l|ll} \toprule
\multirow{2}{*}{\textbf{Model}} & \multirow{2}{*}{\begin{tabular}[c]{@{}l@{}}\textbf{Train Language(s)} \\ \textbf{Configuration} \end{tabular}} & \multicolumn{3}{c|}{ Test on \lareqa{}} & \multirow{2}{*}{\begin{tabular}[c]{@{}l@{}}\textbf{Train Language(s)} \\ \textbf{Configuration} \end{tabular}} & \multicolumn{2}{c}{Test on \stsbm{}} \\ 
& & Multilingual & Bilingual & Monolingual & & Bilingual & Monolingual \\
\midrule
\rowcolor{lightgray}  \multicolumn{8}{c}{Zero-Shot Baselines}   \\ \hline
%\laser{} & - & 13.5 \small{$\pm$ 0.7} & ? & ? & 19.4 \small{$\pm$ 13.3} & 28.9 \small{$\pm$ 9.2}  \\
\labse{} & - & 48.7 \small{$\pm$ 2.6} (6) & 73.0  \small{$\pm$ 1.3} (6) & 77.7 \small{$\pm$ 1.7} (6) & - &  72.3 \small{$\pm$ 7.1} (7) & 77.0 \small{$\pm$ 6.2} (8)   \\

\sbert{} & - & \underline{57.0} \small{$\pm$ 2.9} (4) & \underline{\textit{77.5}} \small{$\pm$ 1.1} (2) & \underline{\textit{80.7}} \small{$\pm$ 1.4} (2) &  - &  \underline{80.2} \small{$\pm$ 5.7} (3) & \underline{82.6} \small{$\pm$ 5.5} (4) \\
\rowcolor{lightgray}  \multicolumn{8}{c}{+Few-Shot Learning}   \\ \hline
\sbert{}+\finetune{} & \monobil{} & 47.0  \small{$\pm$ 4.2} (8) & 68.6 \small{$\pm$ 2.1} (8)  & 71.9 \small{$\pm$ 2.2} (8) & \monobil{} &   77.1 \small{$\pm$ 3.4} (8) & \textit{82.8} \small{$\pm$ 3.1} (2)  \\

\sbert{}+\maml{}(*) & \trans{} & 57.2 \small{$\pm$ 3.5} (3) & 77.1 \small{$\pm$ 1.3} (4) & 80.0 \small{$\pm$ 3.5} (5) & \monobil{} &   \underline{79.9} \small{$\pm$ 2.9} (4) & 82.7 \small{$\pm$ 3.2} (3)   \\

\sbert{}+\alignmaml{}(*) & \monobilmulti{} & \underline{\textit{57.6}} \small{$\pm$ 3.2} (2) & \underline{77.4} \small{$\pm$ 1.5} (3) & \underline{80.6} \small{$\pm$ 1.5} (3) & \monobilmulti{}(**) & 79.5 \small{$\pm$ 2.7}  (5) & \underline{\textbf{85.4} }\small{$\pm$ 1.3} (1) \\
 \rowcolor{lightgray}
\multicolumn{8}{c}{+Machine-Translation}   \\ \hline
%\sbert{}+\translatetest{} & - & \textit{63.5 \small{$\pm$ 0.4}} & %73.9 \small{$\pm$ 1.5}  & 75.5 \small{$\pm$ 1.5}  & - & - &\underline{\textbf{80.9 \small{$\pm$ 6.0}}} & \underline{\textit{84.9 \small{$\pm$ 5.0}}}   \\ \sbert{}+\maml{}+\translatetest{}& Best transfer mode &  \underline{\textbf{64.2 \small{$\pm$ 0.2}}} & - & - & 77.9 \small{$\pm$ 6.6} & 84.1 \small{$\pm$ 6.3} \\ \hline
\multirow{3}{*}{\sbert{}+\translatetrain{}+\finetune{}} & Arabic & \underline{47.3} \small{$\pm$ 3.8} (7) & \underline{69.1} \small{$\pm$ 2.3} (7)  & \underline{72.6} \small{$\pm$ 2.7} (7) & Turkish & \underline{74.3} \small{$\pm$ 6.5} (9) & \underline{81.4} \small{$\pm$ 7.1} (5) \\
& Average over test languages & 46.1 \small{$\pm$ 4.4} (9) & 67.7 \small{$\pm$ 2.7} (9) & 71.1 \small{$\pm$ 3.0} (9) & Average over test languages  & 69.0 \small{$\pm$ 10.7} (10) & 78.4 \small{$\pm$ 9.1} (6) \\ 
 & Jointly over test languages &  45.0 \small{$\pm$ 3.4} (10) & 66.1 \small{$\pm$ 1.7} (10) & 70.6 \small{$\pm$ 2.1} (10) & Jointly over test languages & 68.9 \small{$\pm$ 8.3} (11) & 77.1 \small{$\pm$ 10.1} (7) \\ \cline{2-8}
\multirow{3}{*}{\sbert{}+\translatetrain{}+\maml{}(*)} & Arabic & \underline{\textbf{58.0}} \small{$\pm$ 3.2} (1) & \underline{\textbf{78.1}} \small{$\pm$ 1.4} (1) & \underline{\textbf{81.1}} \small{$\pm$ 1.5} (1) & Turkish & \textit{80.4} \small{$\pm$ 5.7} (2) & \textit{\underline{82.8}} \small{$\pm$ 5.5} (2) \\
& Average over test languages & 57.0 \small{$\pm$ 3.7} (4) & 77.4 \small{$\pm$ 2.1} (3) & 80.4 \small{$\pm$ 2.1} (4) & Average over test languages & 79.1 \small{$\pm$ 5.8} (6) & 82.7 \small{$\pm$ 6.2} (3) \\
& Jointly over test languages & 56.4 \small{$\pm$ 3.7} (5) & 77.0 \small{$\pm$ 1.6} (5) & 80.0 \small{$\pm$ 1.5} (5) & Jointly over test languages & \textbf{\underline{80.5}} \small{$\pm$ 5.7} (1) & 82.6 \small{$\pm$ 5.6} (4) \\ 
\bottomrule
\end{tabular}
}
\caption{\label{tab:avg-all-lareqa-stsb} This is a comparison of different few-shot learning, zero-shot baselines, and machine translation models under a variety of language configuration scenarios. For \lareqa{} and \stsbm{}, we report \map{} and \pearcorr{}, respectively. All results are evaluated over 5-fold cross-validation and averaged over multiple language choices. The same model checkpoint is used for all three task language evaluation variants for each row and dataset (except when the average is reported). \mono, \bil, and \multi stand for monolingual, bilingual, and multilingual semantic search. \trans{} denotes the meta-transfer mode that uses \monobil{} and \bilmulti{} in meta-training and meta-validation, respectively. Models in (*) are our main contribution. (**) means that we use machine-translated data to do that experiment as \stsbm{} is not a parallel corpus. Best and second-best results for each benchmark and evaluation mode are highlighted in \textbf{bold} and \textit{italicized} respectively, whereas the best results across each model category are \underline{underlined}. Ranks from best to worst are given in each model and evaluation mode.\footref{note1}}
\vspace{-0.5cm}
\end{table*}

%  Models, where ranks of at least two different evaluation modes (multilingual, bilingual, and monolingual) are consistent are highlighted in \textbf{bold}.

\section{Results \& Analysis}
\label{sec:results}
\vspace{-0.2cm}
In this section, we present the results obtained using different meta-learning model variants compared to the baselines in multilingual, bilingual, and monolingual task language variants. All experiments are evaluated using 5-fold cross-validation and then the mean and standard deviation are reported. Following XTREME-R~\cite{ruder-etal-2021-xtreme} and SemEval-2017~\cite{cer-etal-2017-semeval}, scores are reported using mean average precision at 20 (\textbf{\map{}}) and Pearson correlation coefficient percentage (\textbf{\pearcorr{}}) for \lareqa{} and \stsbm{}, respectively.  
\vspace{-0.2cm}

\subsection{Multilingual, Bilingual, and Monolingual Performance Evaluation}
\label{subsec:multi-bil-mono-perf}

Table~\ref{tab:avg-all-lareqa-stsb} summarizes multilingual, bilingual, and monolingual performances across different baselines and model variants for both semantic search benchmarks. On average, we notice that \alignmaml{} achieves better results than \maml{} or \sbert{} zero-shot base model and significantly better than \finetune{}. It is worth noting that we report the results for \maml{} using \trans{} mode, which is trained over a combination of \monobil{} and \bilmulti{} in the meta-training and meta-validation stages, respectively. This suggests that \alignmaml{} helps more in bridging the gap between those transfer modes.  We observe that fine-tuning baselines are consistently weak compared to different meta-learning model variants, especially for \lareqa{}. We conjecture that fine-tuning is over-fitting to the small amounts of training data, unlike meta-learning approaches which are more robust against that. However, for \stsbm{}, the gap between fine-tuning and meta-learning while still existing and to the favor of meta-learning is a bit reduced. We hypothesize that even meta-learning models are suffering from meta-overfitting to some degree in this case for \stsbm{}.

% \begin{figure*}[ht!]
%     \centering
%     \includegraphics[width=1\textwidth]{images/languages.pdf}
%      \caption{\label{fig:multi-perf-lareqa-languages-barplots}\map{} and \pearcorr{} 5-fold cross-validated multilingual performance evaluation evaluated on \lareqa{} and \stsbm{} on the first two and last subplots, respectively. The first two subplots on the left show the performance evaluation on Arabic and Russian used in few-shot and zero-shot evaluations, respectively, whereas the last one on the right shows the performance on Spanish-Spanish where Spanish is covered in few-shot learning. There are consistent gains in favor of meta-learning and meta-distillation learning compared to their respective fine-tuning counterparts on top of vanilla \sbert{}.}
% \end{figure*}

\begin{figure*}[ht!]
    \begin{subfigure}[b]{0.5\textwidth}
        \centering
        \includegraphics[width=1\textwidth]{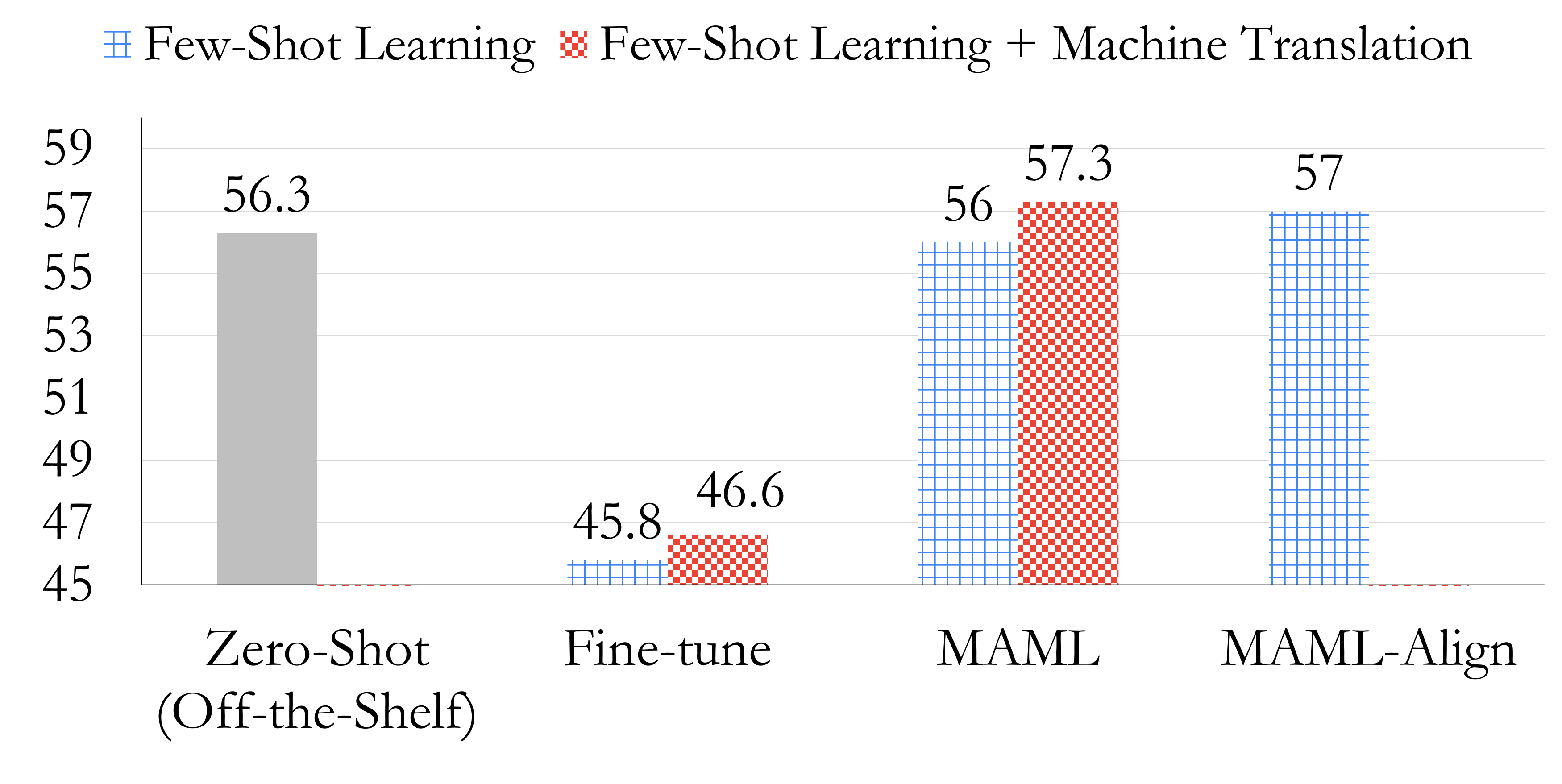}
        \caption{Few-shot multilingual evaluation on Arabic in \lareqa{}.}
        \label{fig:few-shot-arabic}
    \end{subfigure}
    \begin{subfigure}[b]{0.5\textwidth}
        \includegraphics[width=1\textwidth]{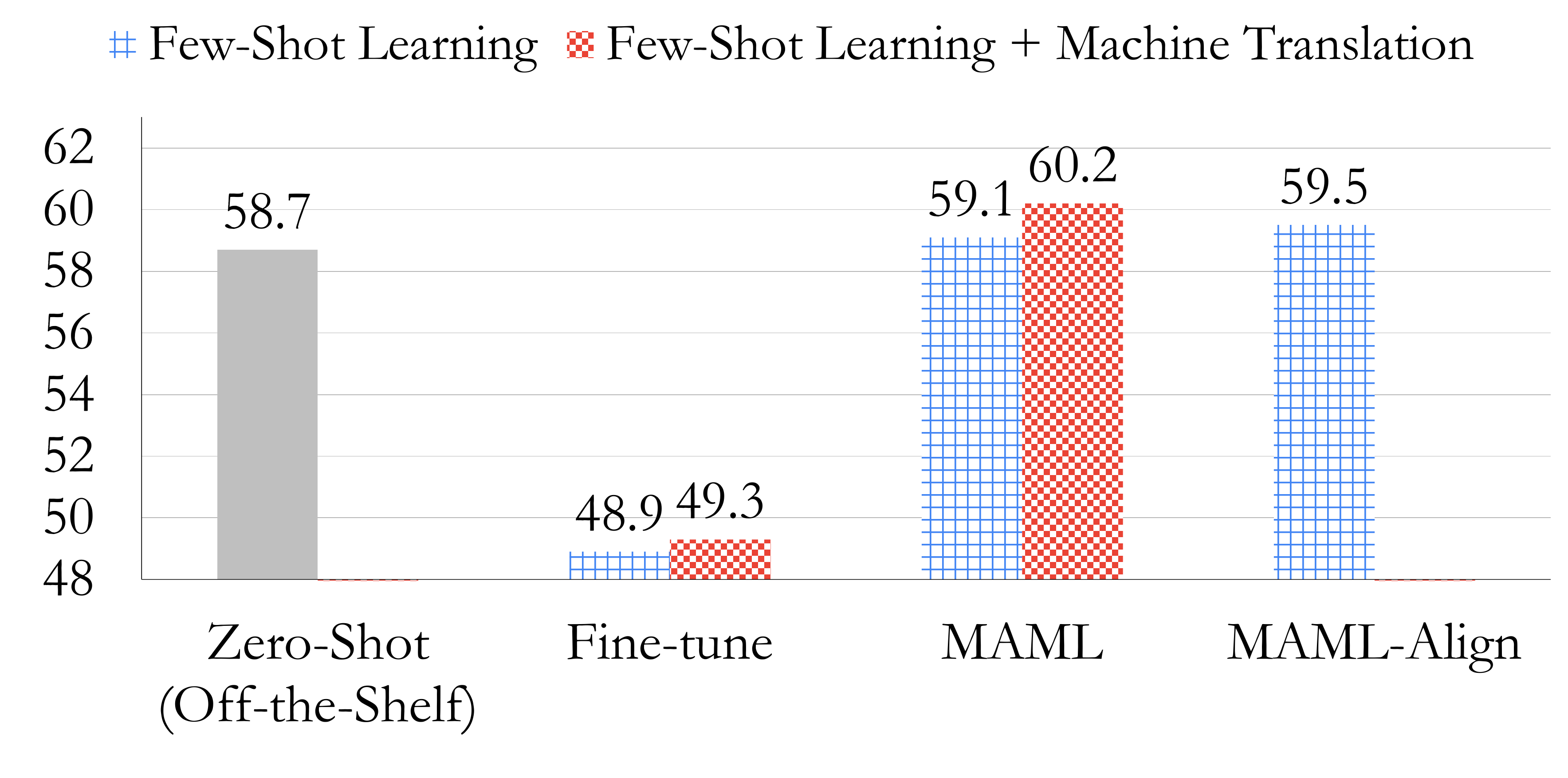}
         \caption{Zero-shot multilingual evaluation on Russian in \lareqa{}.}
        \label{fig:zero-shot-russian}
     \end{subfigure}
     
     % \begin{center}
     % \end{center}
     \begin{subfigure}[b]{0.5\textwidth}
         \includegraphics[width=1\textwidth]{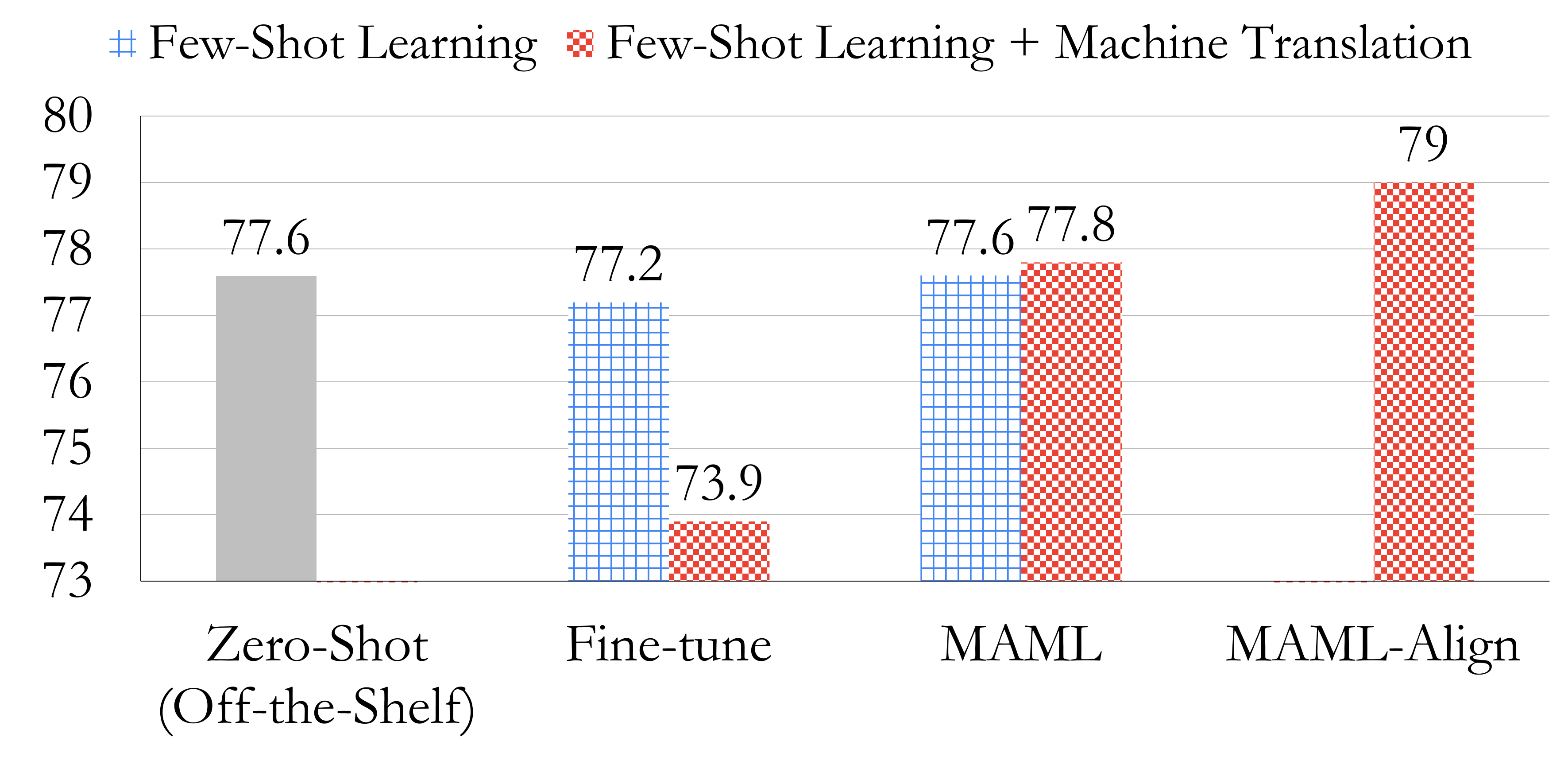}
         \caption{Few-shot evaluation on Arabic-Arabic in \stsbm{}.}
        \label{fig:few-shot-arabic-arabic}
     \end{subfigure}
     \begin{subfigure}[b]{0.5\textwidth}
         \includegraphics[width=1\textwidth]{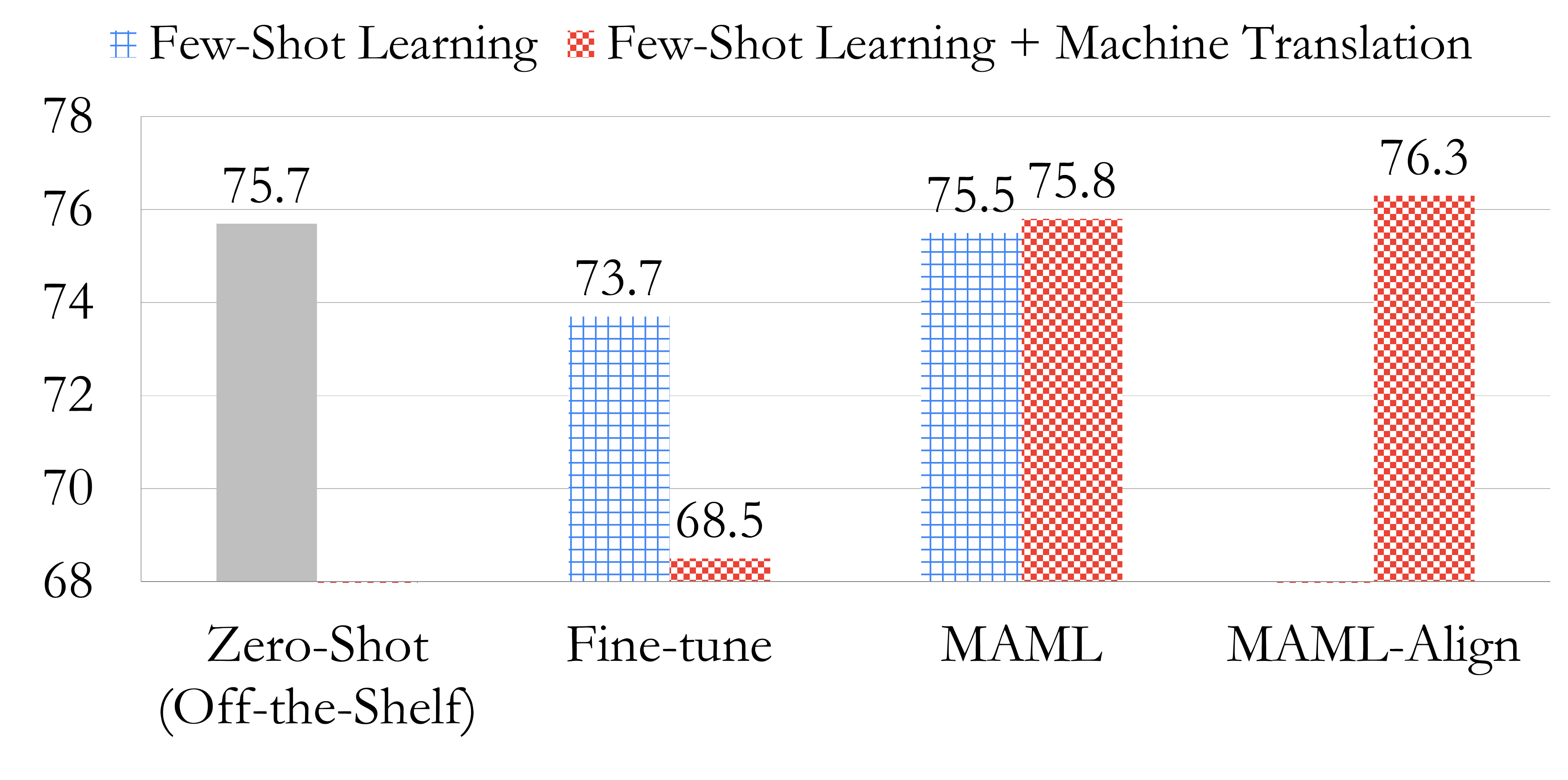}
         \caption{Few-shot evaluation on Turkish-English in \stsbm{}.}
        \label{fig:few-shot-turkish-english}
     \end{subfigure}
 \caption{\label{fig:multi-perf-lareqa-languages-barplots}\map{} and \pearcorr{} 5-fold cross-validated multilingual performance evaluation evaluated on \lareqa{} and \stsbm{} on the first and last two subplots, respectively. The first two subplots show the performance evaluation on Arabic and Russian used in few-shot and zero-shot evaluations, respectively, whereas the two subplots in the second-row showcase monolingual and bilingual performances on Arabic-Arabic and Turkish-English where Arabic, Turkish, and English are all covered in few-shot learning. There are consistent gains in favor of meta-learning and meta-distillation learning compared to their fine-tuning counterparts on top of off-the-shelf model (\sbert{} only) for all types of evaluations.}
 \vspace{-0.5cm}
\end{figure*}

%\meryem{Add color hues for different groups of models zero-shot: \sbert{}, few-shot: (\sbert{}+\finetune{}, \sbert{}+\maml{}), machine translation + few-shot (\sbert{}+\translatetrain{}+\maml{}, \sbert{}+\translatetrain{}+\finetune{}) and machine translation evaluation (\sbert{}+\translatetest{}, \sbert{}+\translatetest{}+\maml{})}

We notice that \maml{} on top of machine-translated data boosts the performance on \lareqa{} in all evaluation task language evaluation variants and reaches the best compromise in terms of multilingual, bilingual, and monolingual performances. At the same time, not all languages used in the machine-translated data provide an equal boost to the performance, as shown by the average performance, due to noisy translations for certain languages. Although there is usually a correlation between different models in terms of their monolingual, bilingual, and multilingual performances, there is a slight drop in the monolingual and bilingual performances for \alignmaml{} compared to the zero-shot baseline. This means that there is still a compromise and gaps between multilingual, monolingual, and bilingual performances. This suggests that we should advocate for a balanced evaluation over different modes to get better insights into which models are more robust and consistent. Figure~\ref{fig:multi-perf-lareqa-languages-barplots} highlights a more fine-grained comparison between different model categories on two languages and language pairs for each benchmark.\footnote{\label{note1}More fine-grained results for all languages and for both benchmarks can be found in Tables~\ref{tab:asym-search-lareqa-languages} and~\ref{tab:sym-search-stsb-languages} in Appendix~\ref{app:more-results}.} We notice that the gain in favor of meta-learning approaches is consistent across different languages and language pairs and also applies to languages used for zero-shot learning. 
\vspace{-0.3cm}
%\maml{} on top of \translatetrain{} improves the generalization even better than \alignmaml{} on the original base data. 

\subsection{Ablation Studies}
\label{sec:ablation-studies}

\begin{figure}[ht]
\centering
\includegraphics[width=0.5\textwidth]%{images/samplingmodes.pdf}
{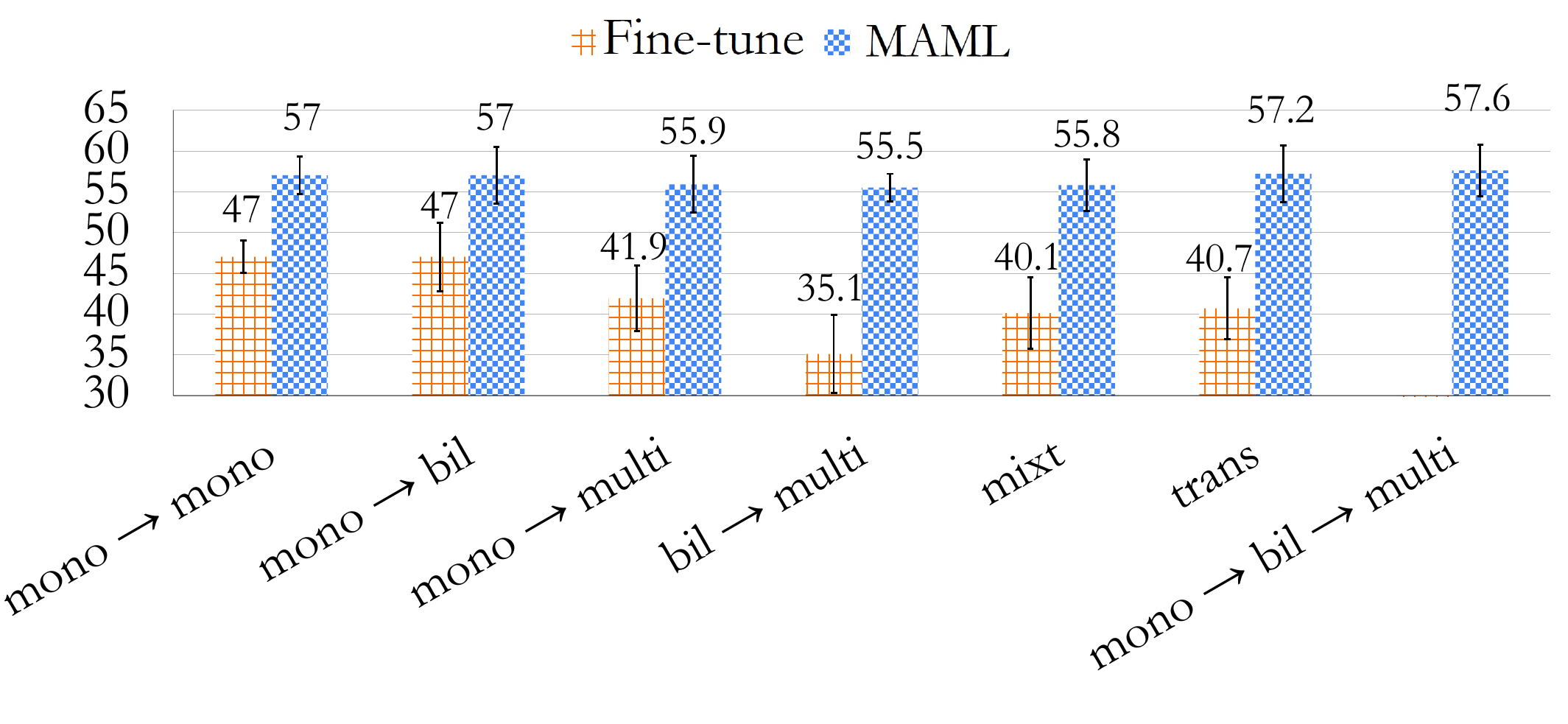}
\vspace{-0.7cm}
\caption{\label{fig:transfer-modes-lareqa} \map{} multilingual 5-fold cross-validated performance on \lareqa{} between different meta-transfer modes for \finetune{} and \maml{} models. The gap is large between \finetune{} and \maml{} across all meta-transfer modes and is even larger to the favor of \maml{} when \trans{} mode (the composed mode that mixes between \monobil{} and \bilmulti{} in the meta-training and meta-validation, respectively) is used.}
\vspace{-0.3cm}
\end{figure}

Due to the lack of parallelism in \stsbm{} making a multilingual evaluation not possible, we focus hereafter on \lareqa{} in the remaining analysis and ablation studies.
Figure~\ref{fig:transfer-modes-lareqa} shows the results across different modes of transfer for \finetune{} and \maml{}. Among all transfer modes, \trans{}, \monobil{}, and \monomono{} have the best gains, whereas \bilmulti{} and \mixt{} are the weakest forms of transfer. \trans{} is the best meta-transfer mode, especially for \maml{} and this suggests that curating different transfer modes for different meta-learning processes is beneficial and leads to better generalization than fine-tuning on them jointly. \mixt{} is weaker than \trans{} and this implies that jointly optimizing different forms of transfers of meta-tasks makes it harder for \maml{} to learn to converge or generalize. \alignmaml{} is shown to be better for combining different optimization objectives.  

% \begin{figure}[ht]
% \centering
% \includegraphics[width=0.5\textwidth]%{images/samplingmodes.pdf}
% {images/figure4.pdf}
% \caption{\label{fig:transfer-modes-lareqa} \map{} multilingual 5-fold cross-validated performance on \lareqa{} tested for different transfer modes.}
% \end{figure}

\begin{figure}[ht]
\centering
\includegraphics[width=0.5\textwidth]
{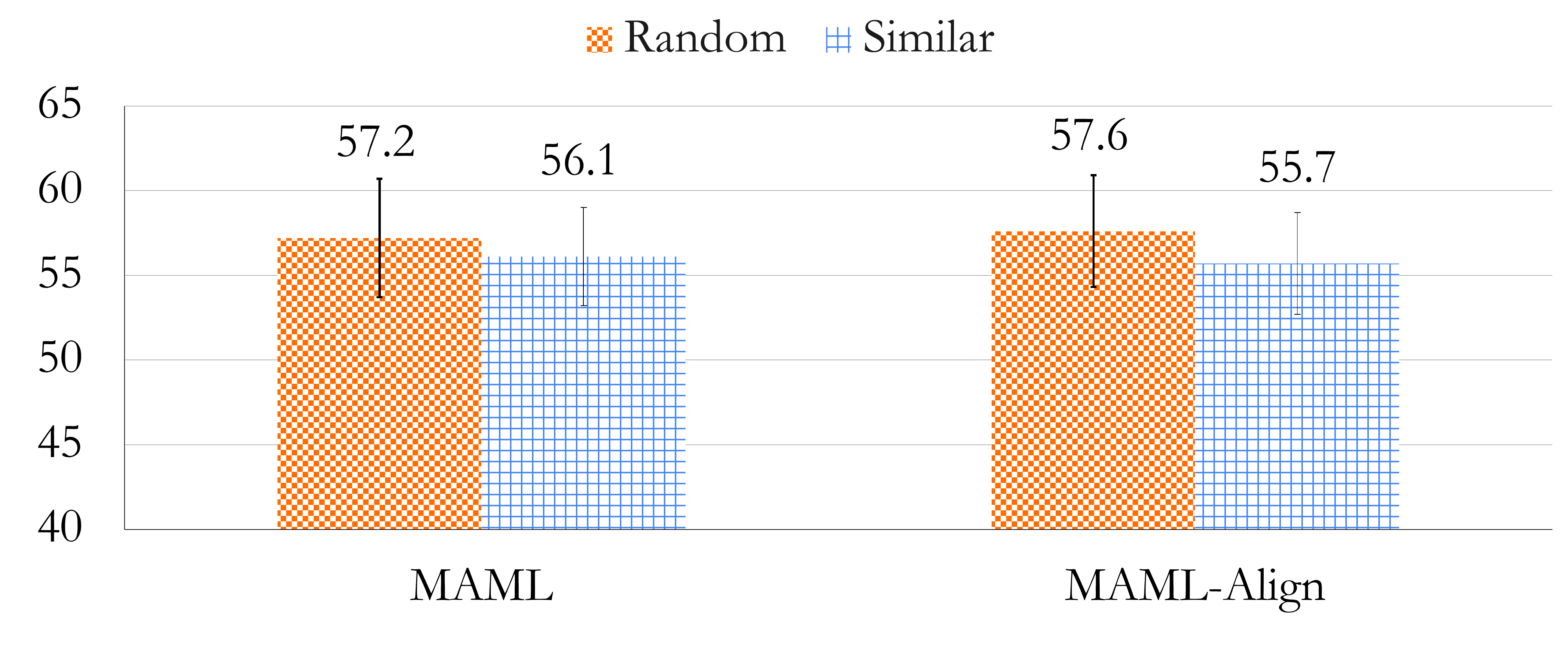}
\caption{\label{fig:rand-sim} \map{} multilingual 5-fold cross-validated performance on \lareqa{} between different query set sampling modes in meta-tasks for \maml{} and \alignmaml{}. We notice that random query sampling has better generalization for both models.}
\vspace{-0.6cm}
\end{figure}

Figure~\ref{fig:rand-sim} shows a multilingual performance comparison between different sampling modes in meta-tasks constructions. In each meta-task, we either sample the query set that is the most similar to its corresponding support set (\textit{Similar}) or randomly (\textit{Random}). We hypothesize that the sampling approach plays a role in stabilizing the convergence and generalization of meta-learning. While we were expecting that sampling for each support set a query set that is the most similar to it would help meta-learning converge faster and thus generalize better, it generalized worse on the multilingual performance in this case. On the other hand, random sampling generalizes better to out-of-sample test distributions leading to lower biases between languages in the multilingual evaluation mode.   %  \maml{} TRANS similar mean_crosslingual: 77.1 1.3 mean_monolingual: 80.2 1.2

\begin{figure}[ht]
\centering
\includegraphics[width=0.5\textwidth]%{images/negative_sampling.pdf}
{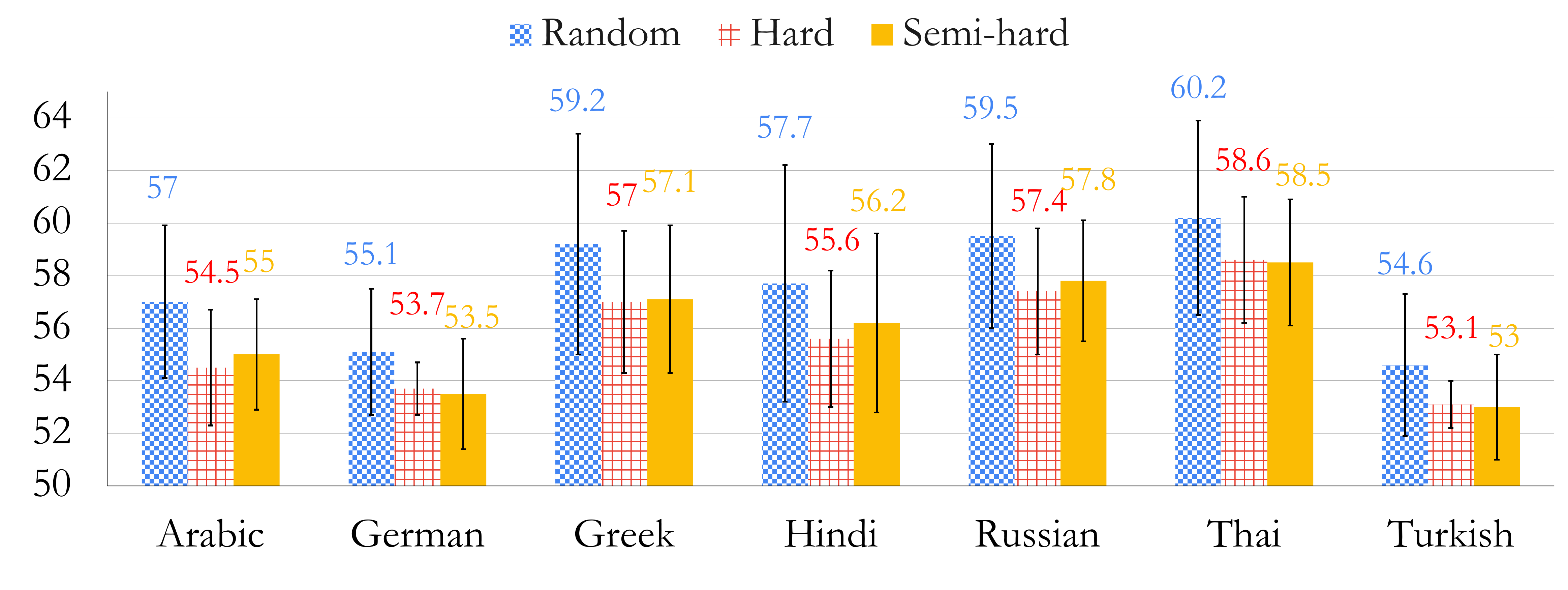}
\caption{\label{fig:triplet-sampling-modes-lareqa} \map{} 5-fold cross-validated multilingual performance over different triplet negative sampling modes on \lareqa{} tested on different languages using \alignmaml{}. We provide both average numbers and standard deviation intervals. Random sampling seems best on average for few-shot learning, whereas hard sampling is more stable across cross-validation splits.}
\vspace{-0.6cm}
\end{figure}

Figure~\ref{fig:triplet-sampling-modes-lareqa} shows the results for different sampling modes of negative examples in the triplet loss. For each support and query set in each meta-task, we either sample random, hard, or semi-hard triplets to test the added value of triplet sampling in few-shot learning. While we expect training with more hard triplets to help converge the triplet loss in \maml{}, the multilingual performance using this type of sampling falls short of random sampling. This is due to the fact that more  sophisticated ways of triplet loss sampling usually require a more careful hyperparameter tuning to pick the right amount of triplets. For few-shot learning applications, this usually results in a significant reduction in the number of training examples, which could further hurt the generalization performance. In future work, we plan to investigate hybrid sampling approaches to monitor at which point in meta-learning the training should focus more on hard or easy triplets. This could be done by proposing a regime for making the sampling of meta-tasks dynamic and flexible to also combat meta-over-fitting.  

%\input{figures/conf_matrices.tex}
% Figures~\ref{fig:x-mono-confusion-matrices} show cross-lingual and monolingual evaluation in the form of confusion matrices. We compare between the S-BERT base model and the best model in T-MAML. The monolingual performances are the diagonal performances and are highlighted in gray whereas the non-diagonal performances are cross-lingual models where the language of the query is strictly different than the language used for the retrieval. Overall, we observe that T-MAML improves the performance monolingually and cross-lingually, there is still a compromise between multilingual, monolingual, and cross-lingual performances.

\section{Conclusion}
\vspace{-0.3cm}
In this work, we adapt multilingual meta-transfer learning combining \maml{} and knowledge distillation to multilingual semantic search. Our experiments show that our multilingual meta-knowledge distillation approach outperforms both vanilla \maml{} and fine-tuning approaches on top of a strong sentence transformers model. We evaluate comprehensively on two types of multilingual semantic search and show improvement over sentence transformers even for languages not covered during meta-learning. 
\section*{Limitations}
Due to the lack of time and resources, exploring different combinations of languages in the construction of the query and the content to be retrieved is not feasible. On top of that, performing extensive hyperparameters search for different model variants, modes of transfer, language combinations, etc is not feasible. We follow a consistent configuration of the hyperparameters for each of the two downstream tasks which we deem to be a fair comparison across all setups, model variants. The insights from this study are tied to the experimental setup that we describe extensively in the main paper and appendix. We also have memory constraints when it comes to training meta-learning algorithms to deal with ranking and retrieval of sentences from multiple languages at the same time for one query. Our memory constraints make it challenging to explore more sophisticated state-of-the-art Sentence Transformers such as sentence-T5 or GPT Sentence Embeddings SGPT~\cite{jianmo-sentT5, Muennighoff-GPTSemSearch-arxiv22}. Applying \maml{} as an upstream model on top of T5-based downstream model makes it even more computationally infeasible. Our main goal is to show the advantage of meta-learning and since our upstream approach is model-agnostic that can be continuously adapted to novel embedding approaches as they evolve. There is also a shortage of large-scale multilingual semantic search datasets, especially for the symmetric case and especially at the phrase level. This makes our evaluation a bit restricted to the bilingual and monolingual for symmetric semantic search. In future work, we plan to construct and annotate semantic search for ambiguous short queries aligned at the multilingual case. 

% \section{Limitations \& Future Work}
% For the remainder of this internship, we will first focus on finishing running the experiments for all \maml{} and \finetune{} baselines and will also report their corresponding monolingual and cross-lingual performance. In parallel to that, we will refine our formulation of the new approach based on meta-distillation learning. We think that the meta-learning direction is very promising and will explore ways to improve the training objective for different transfer modes to come up with a good compromise between monolingual, cross-lingual, and multilingual modes of transfer.

% \section{Things I learned from this Internship}
% This internship has been a great opportunity where I have been fortunate and cannot be grateful enough to my mentors and university supervisor for their support to work on a topic that fall into the direction of my PhD thesis. I have learned a lot professionally and personally. I took part in numerous events such as code quality jam, tech industry transfer, research & PhD intern round-tables, professional development, social hours, wellness, etc. I am also super honored to receive with my team in the code quality jam the prize of the most robust code. I will do my best to apply all my knowledge and research skills in this project and I hope to nurture the collaboration with my Adobe mentors outside of the scope of this summer internship.

\bibliography{custom_refined}

\begin{thebibliography}{53}
\expandafter\ifx\csname natexlab\endcsname\relax\def\natexlab#1{#1}\fi

\bibitem[{Arnold et~al.(2020)Arnold, Mahajan, Datta, Bunner, and
  Zarkias}]{arnold2020learn2learn}
Sébastien M.~R. Arnold, Praateek Mahajan, Debajyoti Datta, Ian Bunner, and
  Konstantinos~Saitas Zarkias. 2020.
\newblock \href {http://arxiv.org/abs/2008.12284} {learn2learn: A library for
  meta-learning research}.

\bibitem[{Artetxe et~al.(2020)Artetxe, Ruder, and
  Yogatama}]{artetxe-etal-2020-cross}
Mikel Artetxe, Sebastian Ruder, and Dani Yogatama. 2020.
\newblock \href {https://doi.org/10.18653/v1/2020.acl-main.421} {On the
  cross-lingual transferability of monolingual representations}.
\newblock In \emph{Proceedings of the 58th Annual Meeting of the Association
  for Computational Linguistics}, pages 4623--4637, Online. Association for
  Computational Linguistics.

\bibitem[{Carvalho et~al.(2008)Carvalho, Elsas, Cohen, and
  Carbonell}]{carvalho2008meta}
Vitor~R Carvalho, Jonathan~L Elsas, William~W Cohen, and Jaime~G Carbonell.
  2008.
\newblock \href
  {https://www.semanticscholar.org/paper/A-Meta-Learning-Approach-for-Robust-Rank-Learning-Carvalho-Elsas/8a94106364576f0aa79dccfb30f0536514408249}
  {A meta-learning approach for robust rank learning}.
\newblock In \emph{SIGIR 2008 workshop on learning to rank for information
  retrieval}, volume~1.

\bibitem[{Cer et~al.(2017)Cer, Diab, Agirre, Lopez-Gazpio, and
  Specia}]{cer-etal-2017-semeval}
Daniel Cer, Mona Diab, Eneko Agirre, I{\~n}igo Lopez-Gazpio, and Lucia Specia.
  2017.
\newblock \href {https://doi.org/10.18653/v1/S17-2001} {{S}em{E}val-2017 task
  1: Semantic textual similarity multilingual and crosslingual focused
  evaluation}.
\newblock In \emph{Proceedings of the 11th International Workshop on Semantic
  Evaluation ({S}em{E}val-2017)}, pages 1--14, Vancouver, Canada. Association
  for Computational Linguistics.

\bibitem[{Chen et~al.(2020)Chen, Hsu, Lee, and
  Lee}]{DBLP:conf/interspeech/ChenHLL20}
Yi{-}Chen Chen, Jui{-}Yang Hsu, Cheng{-}Kuang Lee, and Hung{-}yi Lee. 2020.
\newblock \href {https://doi.org/10.21437/Interspeech.2020-1315} {{DARTS-ASR:}
  differentiable architecture search for multilingual speech recognition and
  adaptation}.
\newblock In \emph{Interspeech 2020, 21st Annual Conference of the
  International Speech Communication Association, Virtual Event, Shanghai,
  China, 25-29 October 2020}, pages 1803--1807. {ISCA}.

\bibitem[{Clark et~al.(2020)Clark, Choi, Collins, Garrette, Kwiatkowski,
  Nikolaev, and Palomaki}]{clark2020tydi}
Jonathan~H. Clark, Eunsol Choi, Michael Collins, Dan Garrette, Tom Kwiatkowski,
  Vitaly Nikolaev, and Jennimaria Palomaki. 2020.
\newblock \href {https://doi.org/10.1162/tacl_a_00317} {{T}y{D}i {QA}: A
  benchmark for information-seeking question answering in typologically diverse
  languages}.
\newblock \emph{Transactions of the Association for Computational Linguistics},
  8:454--470.

\bibitem[{Conneau et~al.(2020)Conneau, Khandelwal, Goyal, Chaudhary, Wenzek,
  Guzm{\'a}n, Grave, Ott, Zettlemoyer, and
  Stoyanov}]{conneau-etal-2020-unsupervised}
Alexis Conneau, Kartikay Khandelwal, Naman Goyal, Vishrav Chaudhary, Guillaume
  Wenzek, Francisco Guzm{\'a}n, Edouard Grave, Myle Ott, Luke Zettlemoyer, and
  Veselin Stoyanov. 2020.
\newblock \href {https://doi.org/10.18653/v1/2020.acl-main.747} {Unsupervised
  cross-lingual representation learning at scale}.
\newblock In \emph{Proceedings of the 58th Annual Meeting of the Association
  for Computational Linguistics}, pages 8440--8451, Online. Association for
  Computational Linguistics.

\bibitem[{Conneau et~al.(2018)Conneau, Rinott, Lample, Williams, Bowman,
  Schwenk, and Stoyanov}]{conneau-etal-2018-xnli}
Alexis Conneau, Ruty Rinott, Guillaume Lample, Adina Williams, Samuel Bowman,
  Holger Schwenk, and Veselin Stoyanov. 2018.
\newblock \href {https://doi.org/10.18653/v1/D18-1269} {{XNLI}: Evaluating
  cross-lingual sentence representations}.
\newblock In \emph{Proceedings of the 2018 Conference on Empirical Methods in
  Natural Language Processing}, pages 2475--2485, Brussels, Belgium.
  Association for Computational Linguistics.

\bibitem[{Devlin et~al.(2019)Devlin, Chang, Lee, and
  Toutanova}]{bert-devlin-naacl19}
Jacob Devlin, Ming-Wei Chang, Kenton Lee, and Kristina Toutanova. 2019.
\newblock \href {https://doi.org/10.18653/v1/N19-1423} {{BERT}: Pre-training of
  deep bidirectional transformers for language understanding}.
\newblock In \emph{Proceedings of the 2019 Conference of the North {A}merican
  Chapter of the Association for Computational Linguistics: Human Language
  Technologies, Volume 1 (Long and Short Papers)}, pages 4171--4186,
  Minneapolis, Minnesota. Association for Computational Linguistics.

\bibitem[{Finn et~al.(2017)Finn, Abbeel, and Levine}]{maml-finn-icml17}
Chelsea Finn, Pieter Abbeel, and Sergey Levine. 2017.
\newblock \href {http://proceedings.mlr.press/v70/finn17a.html} {Model-agnostic
  meta-learning for fast adaptation of deep networks}.
\newblock In \emph{Proceedings of the 34th International Conference on Machine
  Learning, {ICML} 2017, Sydney, NSW, Australia, 6-11 August 2017}, volume~70
  of \emph{Proceedings of Machine Learning Research}, pages 1126--1135. {PMLR}.

\bibitem[{Grefenstette(1998)}]{DBLP:journals/corr/abs-2111-05988}
Gregory Grefenstette. 1998.
\newblock \href {https://aclanthology.org/1998.amta-tutorials.5} {Cross
  language information retrieval}.
\newblock In \emph{Proceedings of the Third Conference of the Association for
  Machine Translation in the Americas: Tutorial Descriptions}, Langhorne, PA,
  USA. Springer.

\bibitem[{Gu et~al.(2018)Gu, Wang, Chen, Li, and
  Cho}]{DBLP:conf/emnlp/GuWCLC18}
Jiatao Gu, Yong Wang, Yun Chen, Victor O.~K. Li, and Kyunghyun Cho. 2018.
\newblock \href {https://doi.org/10.18653/v1/D18-1398} {Meta-learning for
  low-resource neural machine translation}.
\newblock In \emph{Proceedings of the 2018 Conference on Empirical Methods in
  Natural Language Processing}, pages 3622--3631, Brussels, Belgium.
  Association for Computational Linguistics.

\bibitem[{Hoogeveen et~al.(2015)Hoogeveen, Verspoor, and
  Baldwin}]{26-DBLP:conf/adcs/HoogeveenVB15}
Doris Hoogeveen, Karin~M. Verspoor, and Timothy Baldwin. 2015.
\newblock \href {https://doi.org/10.1145/2838931.2838934} {Cqadupstack: {A}
  benchmark data set for community question-answering research}.
\newblock In \emph{Proceedings of the 20th Australasian Document Computing
  Symposium, {ADCS} 2015, Parramatta, NSW, Australia, December 8-9, 2015},
  pages 3:1--3:8. {ACM}.

\bibitem[{Hospedales et~al.(2020)Hospedales, Antoniou, Micaelli, and
  Storkey}]{meta-survey-20}
Timothy~M. Hospedales, Antreas Antoniou, Paul Micaelli, and Amos~J. Storkey.
  2020.
\newblock \href {http://arxiv.org/abs/2004.05439} {Meta-learning in neural
  networks: {A} survey}.
\newblock \emph{CoRR}, abs/2004.05439.

\bibitem[{Hsu et~al.(2020)Hsu, Chen, and Lee}]{DBLP:conf/icassp/HsuCL20}
Jui{-}Yang Hsu, Yuan{-}Jui Chen, and Hung{-}yi Lee. 2020.
\newblock \href {https://doi.org/10.1109/ICASSP40776.2020.9053112} {Meta
  learning for end-to-end low-resource speech recognition}.
\newblock In \emph{2020 {IEEE} International Conference on Acoustics, Speech
  and Signal Processing, {ICASSP} 2020, Barcelona, Spain, May 4-8, 2020}, pages
  7844--7848. {IEEE}.

\bibitem[{Jones et~al.(2008)Jones, Fantino, Newman, and
  Zhang}]{jones-domainspecificquery-ijcnlp08}
Gareth Jones, Fabio Fantino, Eamonn Newman, and Ying Zhang. 2008.
\newblock \href {https://aclanthology.org/I08-6005} {Domain-specific query
  translation for multilingual information access using machine translation
  augmented with dictionaries mined from {W}ikipedia}.
\newblock In \emph{Proceedings of the 2nd workshop on Cross Lingual Information
  Access ({CLIA}) Addressing the Information Need of Multilingual Societies}.

\bibitem[{Laadan et~al.(2019)Laadan, Vainshtein, Curiel, Katz, and
  Rokach}]{laadan2019rankml}
Doron Laadan, Roman Vainshtein, Yarden Curiel, Gilad Katz, and Lior Rokach.
  2019.
\newblock \href {https://arxiv.org/abs/1911.00108} {Rankml: a meta
  learning-based approach for pre-ranking machine learning pipelines}.
\newblock \emph{ArXiv preprint}, abs/1911.00108.

\bibitem[{Langedijk et~al.(2022)Langedijk, Dankers, Lippe, Bos,
  Cardenas~Guevara, Yannakoudakis, and Shutova}]{langedijk2022meta}
Anna Langedijk, Verna Dankers, Phillip Lippe, Sander Bos, Bryan
  Cardenas~Guevara, Helen Yannakoudakis, and Ekaterina Shutova. 2022.
\newblock \href {https://doi.org/10.18653/v1/2022.acl-long.582} {Meta-learning
  for fast cross-lingual adaptation in dependency parsing}.
\newblock In \emph{Proceedings of the 60th Annual Meeting of the Association
  for Computational Linguistics (Volume 1: Long Papers)}, pages 8503--8520,
  Dublin, Ireland. Association for Computational Linguistics.

\bibitem[{Lee et~al.(2022)Lee, Li, and Vu}]{lee-etal-2022-survey}
Hung-yi Lee, Shang-Wen Li, and Thang Vu. 2022.
\newblock \href {https://doi.org/10.18653/v1/2022.naacl-main.49} {Meta learning
  for natural language processing: A survey}.
\newblock In \emph{Proceedings of the 2022 Conference of the North American
  Chapter of the Association for Computational Linguistics: Human Language
  Technologies}, pages 666--684, Seattle, United States. Association for
  Computational Linguistics.

\bibitem[{Lei et~al.(2016)Lei, Joshi, Barzilay, Jaakkola, Tymoshenko,
  Moschitti, and M{\`a}rquez}]{35-DBLP:conf/naacl/LeiJBJTMM16}
Tao Lei, Hrishikesh Joshi, Regina Barzilay, Tommi Jaakkola, Kateryna
  Tymoshenko, Alessandro Moschitti, and Llu{\'\i}s M{\`a}rquez. 2016.
\newblock \href {https://doi.org/10.18653/v1/N16-1153} {Semi-supervised
  question retrieval with gated convolutions}.
\newblock In \emph{Proceedings of the 2016 Conference of the North {A}merican
  Chapter of the Association for Computational Linguistics: Human Language
  Technologies}, pages 1279--1289, San Diego, California. Association for
  Computational Linguistics.

\bibitem[{Lewis et~al.(2020)Lewis, Oguz, Rinott, Riedel, and
  Schwenk}]{lewis-etal-2020-mlqa}
Patrick Lewis, Barlas Oguz, Ruty Rinott, Sebastian Riedel, and Holger Schwenk.
  2020.
\newblock \href {https://doi.org/10.18653/v1/2020.acl-main.653} {{MLQA}:
  Evaluating cross-lingual extractive question answering}.
\newblock In \emph{Proceedings of the 58th Annual Meeting of the Association
  for Computational Linguistics}, pages 7315--7330, Online. Association for
  Computational Linguistics.

\bibitem[{Lin and Chen(2020)}]{lin-chen-2020-preliminary}
Chong-En Lin and Kuan-Yu Chen. 2020.
\newblock \href {https://aclanthology.org/2020.rocling-1.8} {A preliminary
  study on using meta-learning technique for information retrieval}.
\newblock In \emph{Proceedings of the 32nd Conference on Computational
  Linguistics and Speech Processing (ROCLING 2020)}, pages 59--71, Taipei,
  Taiwan. The Association for Computational Linguistics and Chinese Language
  Processing (ACLCLP).

\bibitem[{Litschko et~al.(2022)Litschko, Vulic, Ponzetto, and
  Glavas}]{litschko-multienc-infretr2022}
Robert Litschko, Ivan Vulic, Simone~Paolo Ponzetto, and Goran Glavas. 2022.
\newblock \href {https://doi.org/10.1007/s10791-022-09406-x} {On cross-lingual
  retrieval with multilingual text encoders}.
\newblock \emph{Inf. Retr. J.}, 25(2):149--183.

\bibitem[{Liu et~al.(2022)Liu, Liu, Li, and Liu}]{liu-metakd-arxiv22}
Jihao Liu, Boxiao Liu, Hongsheng Li, and Yu~Liu. 2022.
\newblock \href {https://arxiv.org/abs/2202.07940} {Meta knowledge
  distillation}.
\newblock \emph{ArXiv preprint}, abs/2202.07940.

\bibitem[{Loshchilov and Hutter(2019)}]{DBLP:conf/iclr/LoshchilovH19}
Ilya Loshchilov and Frank Hutter. 2019.
\newblock \href {https://openreview.net/forum?id=Bkg6RiCqY7} {Decoupled weight
  decay regularization}.
\newblock In \emph{7th International Conference on Learning Representations,
  {ICLR} 2019, New Orleans, LA, USA, May 6-9, 2019}. OpenReview.net.

\bibitem[{Lu et~al.(2008)Lu, Xu, and Geva}]{lu-querytrans-aclclp08}
Chengye Lu, Yue Xu, and Shlomo Geva. 2008.
\newblock \href {https://aclanthology.org/O08-3004} {Web-based query
  translation for {E}nglish-{C}hinese {CLIR}}.
\newblock In \emph{International Journal of Computational Linguistics {\&}
  {C}hinese Language Processing, Volume 13, Number 1, March 2008: Special Issue
  on Cross-Lingual Information Retrieval and Question Answering}, pages 61--90.

\bibitem[{M{'}hamdi et~al.(2021)M{'}hamdi, Kim, Dernoncourt, Bui, Ren, and
  May}]{mhamdi-etal-2021-x}
Meryem M{'}hamdi, Doo~Soon Kim, Franck Dernoncourt, Trung Bui, Xiang Ren, and
  Jonathan May. 2021.
\newblock \href {https://doi.org/10.18653/v1/2021.naacl-main.283}
  {{X}-{METRA}-{ADA}: Cross-lingual meta-transfer learning adaptation to
  natural language understanding and question answering}.
\newblock In \emph{Proceedings of the 2021 Conference of the North American
  Chapter of the Association for Computational Linguistics: Human Language
  Technologies}, pages 3617--3632, Online. Association for Computational
  Linguistics.

\bibitem[{Muennighoff(2022)}]{Muennighoff-GPTSemSearch-arxiv22}
Niklas Muennighoff. 2022.
\newblock \href {https://arxiv.org/abs/2202.08904} {{SGPT:} {GPT} sentence
  embeddings for semantic search}.
\newblock \emph{ArXiv preprint}, abs/2202.08904.

\bibitem[{Nayak(2019)}]{blog-nayak-2019}
Pandu Nayak. 2019.
\newblock \href
  {https://blog.google/products/search/search-language-understanding-bert/}
  {Understanding searches better than ever before}.

\bibitem[{Nguyen et~al.(2008)Nguyen, Overwijk, Hauff, Trieschnigg, Hiemstra,
  and de~Jong}]{nguyen-wikiTranslate-clef08}
Dong Nguyen, Arnold Overwijk, Claudia Hauff, Dolf Trieschnigg, Djoerd Hiemstra,
  and Franciska de~Jong. 2008.
\newblock \href {https://doi.org/10.1007/978-3-642-04447-2\_6} {Wikitranslate:
  Query translation for cross-lingual information retrieval using only
  wikipedia}.
\newblock In \emph{Evaluating Systems for Multilingual and Multimodal
  Information Access, 9th Workshop of the Cross-Language Evaluation Forum,
  {CLEF} 2008, Aarhus, Denmark, September 17-19, 2008, Revised Selected
  Papers}, volume 5706 of \emph{Lecture Notes in Computer Science}, pages
  58--65. Springer.

\bibitem[{Ni et~al.(2022)Ni, Hernandez~Abrego, Constant, Ma, Hall, Cer, and
  Yang}]{jianmo-sentT5}
Jianmo Ni, Gustavo Hernandez~Abrego, Noah Constant, Ji~Ma, Keith Hall, Daniel
  Cer, and Yinfei Yang. 2022.
\newblock \href {https://doi.org/10.18653/v1/2022.findings-acl.146}
  {Sentence-t5: Scalable sentence encoders from pre-trained text-to-text
  models}.
\newblock In \emph{Findings of the Association for Computational Linguistics:
  ACL 2022}, pages 1864--1874, Dublin, Ireland. Association for Computational
  Linguistics.

\bibitem[{Nooralahzadeh et~al.(2020)Nooralahzadeh, Bekoulis, Bjerva, and
  Augenstein}]{nooralahzadeh-etal-2020-zero}
Farhad Nooralahzadeh, Giannis Bekoulis, Johannes Bjerva, and Isabelle
  Augenstein. 2020.
\newblock \href {https://doi.org/10.18653/v1/2020.emnlp-main.368} {Zero-shot
  cross-lingual transfer with meta learning}.
\newblock In \emph{Proceedings of the 2020 Conference on Empirical Methods in
  Natural Language Processing (EMNLP)}, pages 4547--4562, Online. Association
  for Computational Linguistics.

\bibitem[{Reimers and Gurevych(2019)}]{reimers-gurevych-2019-sentence}
Nils Reimers and Iryna Gurevych. 2019.
\newblock \href {https://doi.org/10.18653/v1/D19-1410} {Sentence-{BERT}:
  Sentence embeddings using {S}iamese {BERT}-networks}.
\newblock In \emph{Proceedings of the 2019 Conference on Empirical Methods in
  Natural Language Processing and the 9th International Joint Conference on
  Natural Language Processing (EMNLP-IJCNLP)}, pages 3982--3992, Hong Kong,
  China. Association for Computational Linguistics.

\bibitem[{Reimers and Gurevych(2020)}]{reimers-gurevych-2020-making}
Nils Reimers and Iryna Gurevych. 2020.
\newblock \href {https://doi.org/10.18653/v1/2020.emnlp-main.365} {Making
  monolingual sentence embeddings multilingual using knowledge distillation}.
\newblock In \emph{Proceedings of the 2020 Conference on Empirical Methods in
  Natural Language Processing (EMNLP)}, pages 4512--4525, Online. Association
  for Computational Linguistics.

\bibitem[{Robertson and Zaragoza(2009)}]{robertson-zaragoza-bm25-inf-retr}
Stephen~E. Robertson and Hugo Zaragoza. 2009.
\newblock \href {https://doi.org/10.1561/1500000019} {The probabilistic
  relevance framework: {BM25} and beyond}.
\newblock \emph{Found. Trends Inf. Retr.}, 3(4):333--389.

\bibitem[{Roy et~al.(2020)Roy, Constant, Al-Rfou, Barua, Phillips, and
  Yang}]{roy-etal-2020-lareqa}
Uma Roy, Noah Constant, Rami Al-Rfou, Aditya Barua, Aaron Phillips, and Yinfei
  Yang. 2020.
\newblock \href {https://doi.org/10.18653/v1/2020.emnlp-main.477} {{LAR}e{QA}:
  Language-agnostic answer retrieval from a multilingual pool}.
\newblock In \emph{Proceedings of the 2020 Conference on Empirical Methods in
  Natural Language Processing (EMNLP)}, pages 5919--5930, Online. Association
  for Computational Linguistics.

\bibitem[{Ruder et~al.(2021)Ruder, Constant, Botha, Siddhant, Firat, Fu, Liu,
  Hu, Garrette, Neubig, and Johnson}]{ruder-etal-2021-xtreme}
Sebastian Ruder, Noah Constant, Jan Botha, Aditya Siddhant, Orhan Firat, Jinlan
  Fu, Pengfei Liu, Junjie Hu, Dan Garrette, Graham Neubig, and Melvin Johnson.
  2021.
\newblock \href {https://doi.org/10.18653/v1/2021.emnlp-main.802}
  {{XTREME}-{R}: Towards more challenging and nuanced multilingual evaluation}.
\newblock In \emph{Proceedings of the 2021 Conference on Empirical Methods in
  Natural Language Processing}, pages 10215--10245, Online and Punta Cana,
  Dominican Republic. Association for Computational Linguistics.

\bibitem[{Savoy and Braschler(2019)}]{Savoy2019}
Jacques Savoy and Martin Braschler. 2019.
\newblock \href {https://doi.org/10.1007/978-3-030-22948-1_7} {\emph{Lessons
  Learnt from Experiments on the Ad Hoc Multilingual Test Collections at
  CLEF}}, pages 177--200. Springer International Publishing, Cham.

\bibitem[{Schroff et~al.(2015)Schroff, Kalenichenko, and
  Philbin}]{facenet-tripletloss-cvpr2015}
Florian Schroff, Dmitry Kalenichenko, and James Philbin. 2015.
\newblock \href {https://doi.org/10.1109/CVPR.2015.7298682} {Facenet: {A}
  unified embedding for face recognition and clustering}.
\newblock In \emph{{IEEE} Conference on Computer Vision and Pattern
  Recognition, {CVPR} 2015, Boston, MA, USA, June 7-12, 2015}, pages 815--823.
  {IEEE} Computer Society.

\bibitem[{Schuster et~al.(2019)Schuster, Gupta, Shah, and
  Lewis}]{schuster2019cross}
Sebastian Schuster, Sonal Gupta, Rushin Shah, and Mike Lewis. 2019.
\newblock \href {https://doi.org/10.18653/v1/N19-1380} {Cross-lingual transfer
  learning for multilingual task oriented dialog}.
\newblock In \emph{Proceedings of the 2019 Conference of the North {A}merican
  Chapter of the Association for Computational Linguistics: Human Language
  Technologies, Volume 1 (Long and Short Papers)}, pages 3795--3805,
  Minneapolis, Minnesota. Association for Computational Linguistics.

\bibitem[{Tan et~al.(2023)Tan, Heffernan, Schwenk, and
  Koehn}]{tan-multicontrastive-eacl23}
Weiting Tan, Kevin Heffernan, Holger Schwenk, and Philipp Koehn. 2023.
\newblock \href {https://aclanthology.org/2023.eacl-main.108} {Multilingual
  representation distillation with contrastive learning}.
\newblock In \emph{Proceedings of the 17th Conference of the European Chapter
  of the Association for Computational Linguistics}, pages 1477--1490,
  Dubrovnik, Croatia. Association for Computational Linguistics.

\bibitem[{Tarunesh et~al.(2021)Tarunesh, Khyalia, Kumar, Ramakrishnan, and
  Jyothi}]{tarunesh-metamulti-acl21}
Ishan Tarunesh, Sushil Khyalia, Vishwajeet Kumar, Ganesh Ramakrishnan, and
  Preethi Jyothi. 2021.
\newblock \href {https://doi.org/10.18653/v1/2021.eacl-main.314} {Meta-learning
  for effective multi-task and multilingual modelling}.
\newblock In \emph{Proceedings of the 16th Conference of the European Chapter
  of the Association for Computational Linguistics: Main Volume}, pages
  3600--3612, Online. Association for Computational Linguistics.

\bibitem[{van~der Heijden et~al.(2021)van~der Heijden, Yannakoudakis, Mishra,
  and Shutova}]{van2021multilingual}
Niels van~der Heijden, Helen Yannakoudakis, Pushkar Mishra, and Ekaterina
  Shutova. 2021.
\newblock \href {https://doi.org/10.18653/v1/2021.eacl-main.168} {Multilingual
  and cross-lingual document classification: A meta-learning approach}.
\newblock In \emph{Proceedings of the 16th Conference of the European Chapter
  of the Association for Computational Linguistics: Main Volume}, pages
  1966--1976, Online. Association for Computational Linguistics.

\bibitem[{Vaswani et~al.(2017)Vaswani, Shazeer, Parmar, Uszkoreit, Jones,
  Gomez, Kaiser, and Polosukhin}]{DBLP:conf/nips/VaswaniSPUJGKP17}
Ashish Vaswani, Noam Shazeer, Niki Parmar, Jakob Uszkoreit, Llion Jones,
  Aidan~N. Gomez, Lukasz Kaiser, and Illia Polosukhin. 2017.
\newblock \href
  {https://proceedings.neurips.cc/paper/2017/hash/3f5ee243547dee91fbd053c1c4a845aa-Abstract.html}
  {Attention is all you need}.
\newblock In \emph{Advances in Neural Information Processing Systems 30: Annual
  Conference on Neural Information Processing Systems 2017, December 4-9, 2017,
  Long Beach, CA, {USA}}, pages 5998--6008.

\bibitem[{Winata et~al.(2020)Winata, Cahyawijaya, Lin, Liu, Xu, and
  Fung}]{DBLP:conf/acl/WinataCLLXF20}
Genta~Indra Winata, Samuel Cahyawijaya, Zhaojiang Lin, Zihan Liu, Peng Xu, and
  Pascale Fung. 2020.
\newblock \href {https://doi.org/10.18653/v1/2020.acl-main.348} {Meta-transfer
  learning for code-switched speech recognition}.
\newblock In \emph{Proceedings of the 58th Annual Meeting of the Association
  for Computational Linguistics}, pages 3770--3776, Online. Association for
  Computational Linguistics.

\bibitem[{Xiao et~al.(2021)Xiao, Gong, Zhou, Zheng, Liang, and
  Lin}]{DBLP:conf/aaai/XiaoGZZLL21}
Yubei Xiao, Ke~Gong, Pan Zhou, Guolin Zheng, Xiaodan Liang, and Liang Lin.
  2021.
\newblock \href {https://ojs.aaai.org/index.php/AAAI/article/view/17661}
  {Adversarial meta sampling for multilingual low-resource speech recognition}.
\newblock In \emph{Thirty-Fifth {AAAI} Conference on Artificial Intelligence,
  {AAAI} 2021, Thirty-Third Conference on Innovative Applications of Artificial
  Intelligence, {IAAI} 2021, The Eleventh Symposium on Educational Advances in
  Artificial Intelligence, {EAAI} 2021, Virtual Event, February 2-9, 2021},
  pages 14112--14120. {AAAI} Press.

\bibitem[{Xu et~al.(2021)Xu, Ebner, Yarmohammadi, White, Van~Durme, and
  Murray}]{xu-gradualtuning-aclwork2021}
Haoran Xu, Seth Ebner, Mahsa Yarmohammadi, Aaron~Steven White, Benjamin
  Van~Durme, and Kenton Murray. 2021.
\newblock \href {https://aclanthology.org/2021.adaptnlp-1.22} {Gradual
  fine-tuning for low-resource domain adaptation}.
\newblock In \emph{Proceedings of the Second Workshop on Domain Adaptation for
  NLP}, pages 214--221, Kyiv, Ukraine. Association for Computational
  Linguistics.

\bibitem[{Yang et~al.(2020)Yang, Cer, Ahmad, Guo, Law, Constant,
  Hernandez~Abrego, Yuan, Tar, Sung, Strope, and
  Kurzweil}]{76-DBLP:conf/acl/YangCAGLCAYTSSK20}
Yinfei Yang, Daniel Cer, Amin Ahmad, Mandy Guo, Jax Law, Noah Constant, Gustavo
  Hernandez~Abrego, Steve Yuan, Chris Tar, Yun-hsuan Sung, Brian Strope, and
  Ray Kurzweil. 2020.
\newblock \href {https://doi.org/10.18653/v1/2020.acl-demos.12} {Multilingual
  universal sentence encoder for semantic retrieval}.
\newblock In \emph{Proceedings of the 58th Annual Meeting of the Association
  for Computational Linguistics: System Demonstrations}, pages 87--94, Online.
  Association for Computational Linguistics.

\bibitem[{Zhang et~al.(2020)Zhang, Wang, and Gai}]{zhang-kdf4maml-ecai20}
Min Zhang, Donglin Wang, and Sibo Gai. 2020.
\newblock \href {https://doi.org/10.3233/FAIA200239} {Knowledge distillation
  for model-agnostic meta-learning}.
\newblock In \emph{{ECAI} 2020 - 24th European Conference on Artificial
  Intelligence, 29 August-8 September 2020, Santiago de Compostela, Spain,
  August 29 - September 8, 2020 - Including 10th Conference on Prestigious
  Applications of Artificial Intelligence {(PAIS} 2020)}, volume 325 of
  \emph{Frontiers in Artificial Intelligence and Applications}, pages
  1355--1362. {IOS} Press.

\bibitem[{Zhou et~al.(2022)Zhou, Xu, and McAuley}]{zhou-bertlearns2teach-acl22}
Wangchunshu Zhou, Canwen Xu, and Julian McAuley. 2022.
\newblock \href {https://doi.org/10.18653/v1/2022.acl-long.485} {{BERT} learns
  to teach: Knowledge distillation with meta learning}.
\newblock In \emph{Proceedings of the 60th Annual Meeting of the Association
  for Computational Linguistics (Volume 1: Long Papers)}, pages 7037--7049,
  Dublin, Ireland. Association for Computational Linguistics.

\bibitem[{Zhu et~al.(2021)Zhu, Li, Li, and Odoula}]{zhu-li-li-odoula-2021}
Jeffrey Zhu, Mingqin Li, Jason Li, and Cassandra Odoula. 2021.
\newblock \href
  {https://blogs.bing.com/Engineering-Blog/october-2021/Bing-delivers-more-contextualized-search-using-quantized-transformer-inference-on-NVIDIA-GPUs-in-Azu}
  {Bing delivers more contextualized search using quantized transformer
  inference on nvidia gpus in azure}.

\bibitem[{Ziemski et~al.(2016)Ziemski, Junczys-Dowmunt, and
  Pouliquen}]{85-DBLP:conf/lrec/ZiemskiJP16}
Micha{\l} Ziemski, Marcin Junczys-Dowmunt, and Bruno Pouliquen. 2016.
\newblock \href {https://aclanthology.org/L16-1561} {The {U}nited {N}ations
  parallel corpus v1.0}.
\newblock In \emph{Proceedings of the Tenth International Conference on
  Language Resources and Evaluation ({LREC}'16)}, pages 3530--3534,
  Portoro{\v{z}}, Slovenia. European Language Resources Association (ELRA).

\bibitem[{Zweigenbaum et~al.(2018)Zweigenbaum, Sharoff, and
  Rapp}]{86-DBLP:conf/lrec/ZweigenbaumSR18a}
Pierre Zweigenbaum, Serge Sharoff, and Reinhard Rapp. 2018.
\newblock \href
  {http://lrec-conf.org/workshops/lrec2018/W8/summaries/12\_W8.html} {Overview
  of the third {BUCC} shared task: Spotting parallel sentences in comparable
  corpora}.
\newblock In \emph{11th Workshop on Building and Using Comparable Corpora -
  Special Topic: Comparable Corpora for Asian Languages, BUCC@LREC 2018,
  Miyazaki, Japan, May 8, 2018}. European Language Resources Association.

\end{thebibliography}
\bibliographystyle{acl_natbib}

\appendix
\section{More Related Work}
\label{app:more-related-work}

Given the scarcity of research on multilingual semantic search using meta-learning and knowledge distillation, we analyze independently previous work in the area of semantic search in general, multilingual semantic search, cross-lingual meta-transfer learning, and meta-distillation learning before delving into some applications of meta-transfer learning for retrieval ranking and how our work applies meta-transfer and meta-distillation for multilingual semantic search as a whole. 

\paragraph{Textual Semantic Search}
Textual semantic search is the task of retrieving semantically relevant content for a given query. Unlike traditional keyword-matching information retrieval, semantic search seeks to improve search accuracy by understanding the searcher's intent and disambiguating the contextual meaning of the terms in the query~\cite{Muennighoff-GPTSemSearch-arxiv22}. Semantic search has broad applications in search engines such as Google~\cite{blog-nayak-2019}, Bing~\cite{zhu-li-li-odoula-2021}, etc. They rely on Transformers~\cite{DBLP:conf/nips/VaswaniSPUJGKP17} as their dominant architecture going beyond non-semantic models such as BM25~\cite{robertson-zaragoza-bm25-inf-retr}. 

\paragraph{Multilingual Semantic Search}

Previous work which extends semantic search to different languages is often focused on cross-lingual information retrieval. Progress in cross-lingual information retrieval (CLIR) or semantic search has seen multiple waves~\cite{DBLP:journals/corr/abs-2111-05988}. Traditionally, when we think of CLIR we automatically think of machine translation (MT) as if they are two faces to the same coin. The only difference is that translation tools are used to render documents readable in the case of MT whereas CLIR focuses on rendering them searchable if at the very core translation technology is what is used for CLIR and MT rather than other paradigms such as transfer learning.  Most approaches that fall into this category translate queries into the language of the documents and then perform monolingual search~\cite{lu-querytrans-aclclp08,nguyen-wikiTranslate-clef08,jones-domainspecificquery-ijcnlp08}. While this is an efficient option, that might not be the most effective approach as queries can be so short and ungrammatical making them hard to translate accurately. So, in this case, translating all documents or sentences to the target languages can be used leading to better accuracy but less efficiency. This translation form is even more inefficient in the case of multilingual semantic search where the number of possible language combinations that can be used in the source and target languages can grow exponentially.  Those pipeline approaches suffer from error propagation of the machine translation component into the downstream semantic search, especially for low-resource languages. 

More prominent approaches include transfer learning where both query and documents or sentences are encoded into a shared space. The first class of approaches in this category use pre-trained language models where both the query and the documents are encoded into a shared space. The cross-lingual ability of models like \mbert{} and \xlm{} has been analyzed for different retrieval-based downstream applications including question-answer retrieval~\cite{76-DBLP:conf/acl/YangCAGLCAYTSSK20}, bitext mining~\cite{85-DBLP:conf/lrec/ZiemskiJP16, 86-DBLP:conf/lrec/ZweigenbaumSR18a}, and semantic textual similarity~\cite{26-DBLP:conf/adcs/HoogeveenVB15, 35-DBLP:conf/naacl/LeiJBJTMM16}. \citet{litschko-multienc-infretr2022} systematic empirical study focused on the suitability of SOTA multilingual encoders for cross-lingual document and sentence retrieval tasks across a number of diverse language pairs. They benchmark the performance in unsupervised ad-hoc (setup with no relevance judgments for IR-specific fine-tuning) and supervised sentence and document-level CLIR. In other words, they profile the suitability of SOTA pre-trained multilingual encoders for different CLIR tasks and diverse language pairs across unsupervised, supervised and transfer setups. They also propose localized relevance matching for document-level CLIR (independently score a query against document). For unsupervised document-level CLIR, they show that pre-trained multilingual encoders on average fail to significantly outperform earlier models based on CLWEs. They also show that the performance of those multilingual encoders crucially depends on how one encodes semantic information with the models (treating them as sentence/document encoders directly versus averaging over constituent words and/or subwords). Multilingual sentence encoders fine-tuned on labeled data from sentence pair tasks like natural language inference or semantic text similarity as well as using parallel sentences on the other hand are shown to substantially outperform general-purpose models in sentence-level CLIR. The second class focuses on training training models with information retrieval objectives but it is not clear how they generalize to new languages. In our work, we investigate ways to further improve the transfer of these off-purpose sentences on top of semantic specialization in a data-efficient manner. 

%% TWO PARAGRAPHS 
% Summary of MT related work (one or two paragraphs) and their gains and downsides
% You: \meryem{TODO describe shortcomings of translation} \meryem{TODO add citations in this section and organize it into paragraphs}

% Summary of Cross-lingual transfer learning for semantic search and information retrieval

%%% CL-ReLKT: Cross-lingual Language Knowledge Transfer for Multilingual Retrieval Question Answering

\paragraph{Multilingual Meta-Transfer Learning}

Meta-learning has gained the attention of the NLP community recently with applications in cross-domain, cross-problem, and cross-lingual transfer learning~\cite{lee-etal-2022-survey}. Meta-learning has been leveraged for semantic search-related tasks but only monolingually. \citet{lin-chen-2020-preliminary} is the first work of its kind to device a meta-learning algorithm for information retrieval tasks. They leverage model-agnostic meta-learner (MAML) to learn an initialization of model parameters for the re-ranker of documents by reformulating the problem as a N-way K-Shot setup where query is a category and the document corresponding to it as a positive example and four documents not related to the query. They show that their approach improves over baselines involving vanilla DSSM and Vector Space Models. They also show that fine-tuning in addition to meta-learning lead to more gains. However, they use meta-learning just at the level of the ranker and not for other components like searcher in which they only use traditional approaches like Match 25 to calculate the relationship between query documents and retrieval documents. It is not clear whether meta-learning can be used more in an end-to-end fashion or to improve other components. Other meta-learning work which focus on the re-ranking component include~\citet{laadan2019rankml,carvalho2008meta} but they all follow a pipelined approach.

%\paragraph{Meta-learning for Ranking}

Since there is no prior work leveraging meta-learning for cross-lingual or multilingual semantic search, to the best of our knowledge, we describe in this section. 
The first work of its kind using meta-learning for cross-lingual transfer learning is~\citet{DBLP:conf/emnlp/GuWCLC18}, which is applied to neural machine translation. They extend MAML\cite{maml-finn-icml17} to transfer from multilingual high-resource language tasks to to low-resource languages. They show the competitive advantages of cross-lingual meta-transfer learning compared to other multilingual baselines. Other applications include speech recognition~\cite{DBLP:conf/icassp/HsuCL20, DBLP:conf/acl/WinataCLLXF20, DBLP:conf/interspeech/ChenHLL20, DBLP:conf/aaai/XiaoGZZLL21}, Natural Language Inference(XNLI)~\cite{conneau-etal-2018-xnli} and Multilingual Question Answering(MLQA)~\cite{lewis-etal-2020-mlqa} using X-MAML~\cite{nooralahzadeh-etal-2020-zero}, task-oriented dialog~\cite{schuster2019cross} and TyDiQA~\cite{clark2020tydi} using X-METRA-ADA~\cite{mhamdi-etal-2021-x}, dependency parsing~\cite{langedijk2022meta}. 

%%% cite cross-task cross-lingual transfer
%%% Cross-Lingual Transfer with MAML on Trees
%%% Soft Layer Selection with Meta-Learning for Zero-Shot Cross-Lingual Transfer
%%% Meta-Learning for Fast Cross-Lingual Adaptation in Dependency Parsing
%%% Meta-XNLG: A Meta-Learning Approach Based on Language Clustering for Zero-Shot Cross-Lingual Transfer and Generation
%%% Meta-Learning a Cross-lingual Manifold for Semantic Parsing
%%% Meta-ED: Cross-lingual Event Detection using Meta-learning for Indian Languages

%% cross-lingual meta-learning in general take it from your NAACL paper

%% multilingual meta-learning: describe that one model for multiple languages is more challenging and less explored 
Most recent work adapting meta-learning to applications involving different languages focus on cross-lingual meta-learning. Multilingual meta-learning differs from cross-lingual meta-transfer learning in its support for multiple languages jointly. \citeauthor{mhamdi-etal-2021-x}, for example, propose X-METRA-ADA which performs few-shot learning on one single target language at a time and also enable zero-shot learning on target languages not seen during meta-training or meta-adaptation. Their approach shows gains compared to naive fine-tuning in the few-shot more than the zero-shot learning scenario.~\citet{tarunesh-metamulti-acl21} propose a meta-learning framework for both multi-task and multi-lingual transfer leveraging heuristic sampling approaches. They show that a joint approach to multi-task and multilingual learning using meta-learning enables effective sharing of parameters across multiple tasks and multiple languages thus benefits deeper semantic analysis tasks such as QA, PAWS, NLI, etc.~\citet{van2021multilingual} propose a meta-learning framework and show its effectiveness in both the cross-lingual and multilingual training adaptation settings of document classification. However, their multilingual evaluation is focused on the scenario where the same target languages during meta-testing can be also used as auxiliary languages during meta-training. This motivates us to investigate in this paper more in the direction of multilingual meta-transfer learning, where we test the generalizability of our meta-learning model when it is learned by taking into consideration multiple languages jointly for semantic search.

%%% Meta-Learning for Effective Multi-task and Multilingual Modelling
%%% Multilingual and cross-lingual document classification: a meta-learning approach
%%% "Diversity and Uncertainty in Moderation" are the Key to Data Selection for Multilingual Few-shot Transfer

\paragraph{Meta-Distillation Learning}

Previous works at the intersection of meta-learning and knowledge distillation either use meta-learning as a more effective alternative to the more traditional knowledge distillation methods. Recently, more work has started adopting a meta-learning approach to knowledge distillation by consolidating a feedback loop between the teacher and the student networks where the teacher can learn to better transfer knowledge to the student network~\cite{zhou-bertlearns2teach-acl22} or by meta-learning the distillation hyperparameter tuning~\cite{liu-metakd-arxiv22}. Knowledge distillation has also been leveraged to enhance the portability of MAML networks~\cite{zhang-kdf4maml-ecai20}. It has been shown that a portable MAML with a smaller capacity can further boost few-shot learning better than vanilla MAML. To the best of our knowledge, we are the first to explore knowledge distillation to bridge the gap between different cross-lingual meta-transfer learning models and to enhance the alignment between them.

%% Describe ACL papers in the area
%%% MetaTS: Meta Teacher-Student Network for Multilingual Sequence Labeling with Minimal Supervision
%%% ACL 2022 papers on meta-distil in general

%% need to read more about that

%% how your paper combines all of that

\section{More Details on Base Models}
\label{app:base-models}

\begin{figure}[ht]
\centering
\includegraphics[width=0.45\textwidth]{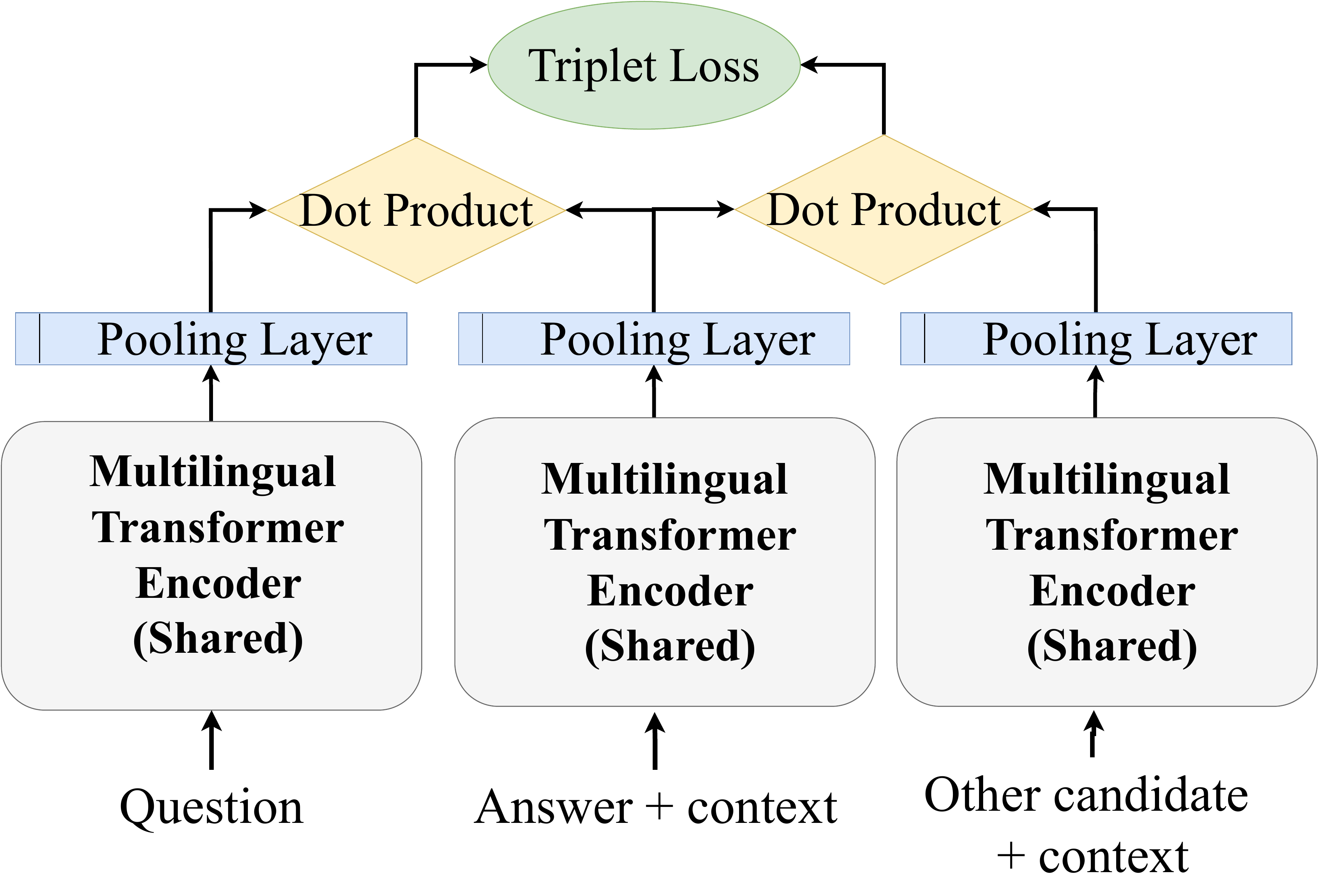}
\caption{\label{fig:asymmetric-base-model} Architecture of Transformer-based triplet encoder for asymmetric semantic search.}
\end{figure}

For asymmetric semantic search, we use a Transformer-based triplet-encoder model. In the original paper on the asymmetric benchmark we evaluate on~\cite{roy-etal-2020-lareqa}, a dual-encoder model is trained using contrastive loss in the form of an in-batch sampled softmax loss. This format reuses for each question answers from other questions in the same batch (batched randomly) as negative examples. Instead, we use triplet loss~\cite{facenet-tripletloss-cvpr2015}, which was also shown to outperform contrastive loss in general. Triplet loss is shown to surpass contrastive loss in general.\footnote{As posited in \url{https://shorturl.at/ktvx9}.} Its strength derives not just from the nature of its function but also from its sampling procedure. This sampling procedure which merely requires positive instances to be closer to negative instances doesn't require gathering as many positive examples as contrastive loss requires. This makes triplet loss more practical in our few-shot learning multilingual/cross-lingual scenario, as it provides more freedom in terms of constructing negative candidates to tweak different sampling techniques from different languages. We thus define a \emph{triplet encoder model} (shown in Figure~\ref{fig:asymmetric-base-model}) with three towers encoding the question, its answer combined with its context, and the negative candidates and their contexts. While those towers are encoded separately, they still share the same Transformer encoder model which is initialized with pre-trained Sentence Transformers. On top of that, two dot products $d(q,p)$ and $d(q,n)$ are computed. $d(q,p)$ is the dot product between the question $q$ and its answer $p$, whereas $d(q,n)$ is between $q$ and its non-answer candidate. Triplet loss is computed as : $\mathcal{L} = \max_{}{(d(q,p) - d(q,n) + margin, 0)}$
% \begin{equation}
% ,
% \end{equation} 
where $margin$ is a tun-able hyperparameter to eventually make each triplet an easy one by pushing the distance $d(a,p)$ closer to 0 and $d(a,n)$ to $d(a,p) + margin$.

Triplets $(q,p,n)$ can be sampled with different levels of difficulty, as follows:
\vspace{-0.2cm}
\begin{itemize}[leftmargin=*]
    \itemsep0em
    \item{\textbf{Easy triplets:} $d(q,p) + margin < d(q,n)$.}
    \item{\textbf{Hard triplets:} $d(q,n) < d(q,p)$.}
    \item{\textbf{Semi-hard triplets:} $d(q,p) < d(q,n) < d(q,p) + margin$.}
\end{itemize}

\begin{figure}[t]
\centering
\includegraphics[width=0.3\textwidth]{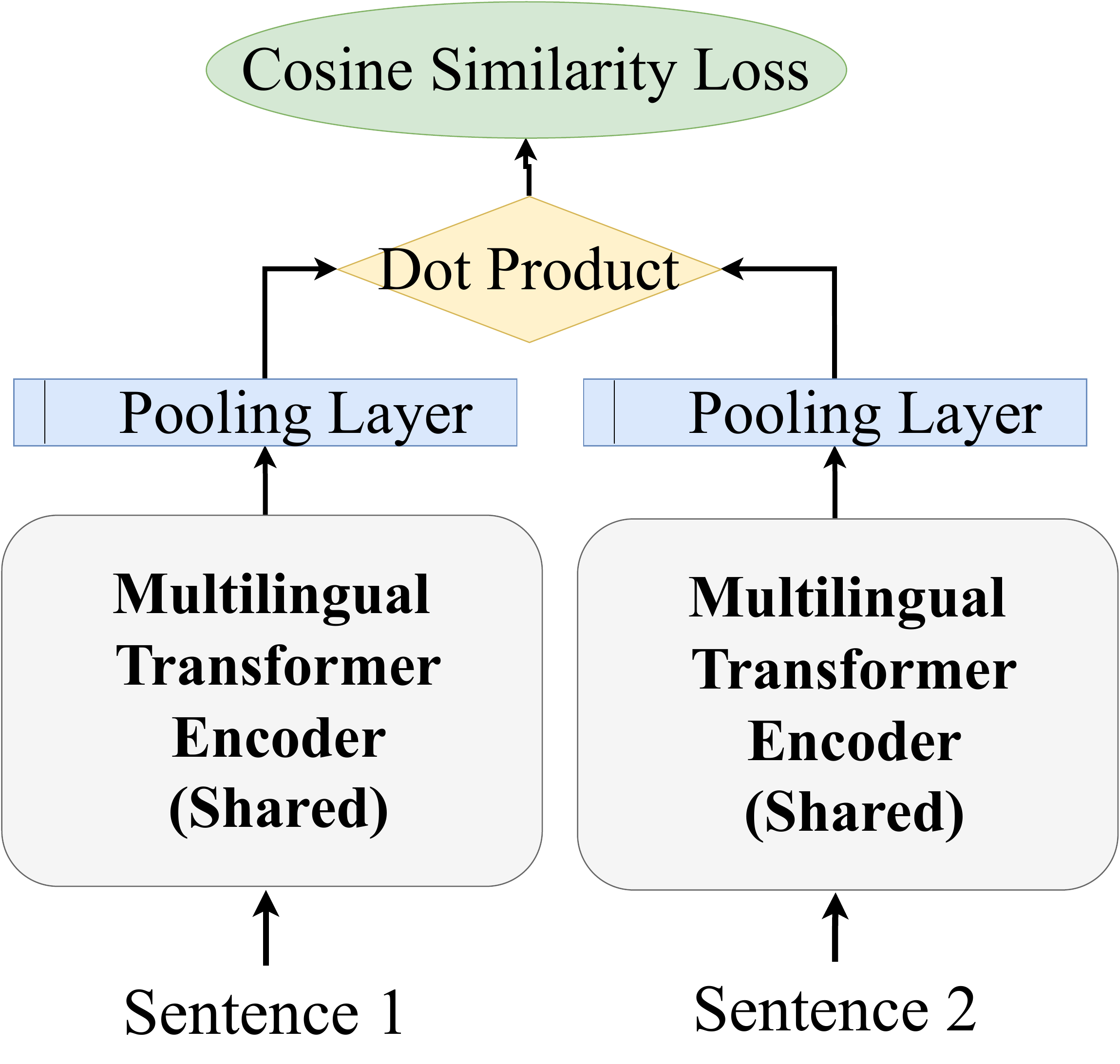}
\caption{\label{fig:symmetric-base-model} Architecture of Transformer-based dual-encoder for symmetric semantic search.}
\end{figure}

For symmetric search, we use a Transformer-based dual-encoder model (shown in Figure~\ref{fig:symmetric-base-model}), which encodes sentence 1 and sentence 2 in each sentence pair separately using the same shared encoder. Then, the cosine similarity score is computed for each sentence pair and the mean squared error (squared L2 norm) is computed between that and the golden score. This is not a retrieval-based task, but a semantic similarity task.

\section{More Experimental Setup Details}
\label{app:more-experimental-setup}

\subsection{Downstream Datasets}
\label{app:datasets}

Tables~\ref{tab:lareqa-stats} and~\ref{tab:stsbmulti-stats} show a summary of the statistics of \lareqa{} and \stsbm{} per language and split, respectively. \xquadr{} in \lareqa{} has been distributed under the CC BY-SA 4.0 license, whereas \stsbm{} has been released under the Creative Commons Attribution-ShareAlike 4.0 International License. The translated datasets from \ensquad{} and \stsben{} are shared under the same license as the original
datasets. \ensquad{} is shared under XTREME benchmark Apache License Version 2.0. \stsben{} scores are under  Creative Commons Attribution-ShareAlike 3.0 Unported (CC BY-SA 3.0) and sentence pairs are shared under Commons Attribution - Share Alike 4.0 International License).

\begin{table}[ht]
\centering
\scalebox{0.95}{
\small
\begin{tabular}{l|l|ll|ll|ll}
\toprule
\multirow{2}{*}{\textbf{Language}} & \multirow{2}{*}{\textbf{ISO}} & \multicolumn{2}{c|}{\textbf{Train}}  & \multicolumn{2}{c|}{\textbf{Dev}} & \multicolumn{2}{c}{\textbf{Test}} \\ 
  &   &  \#Q & \#C & \#Q & \#C & \#Q & \#C \\
\toprule
Arabic & AR & 696 & 783 & 220 & 255 & 274 & 184 \\
German & DE & 696 & 812 & 220 & 256 & 274 & 208 \\
Greek & EL & 696 & 788 & 220 & 254 & 274 & 192 \\
Hindi & HI & 696 & 808 & 220 & 252 & 274 & 184 \\
Russian & RU & 696 & 774 & 220 & 262 & 274 & 183 \\ 
Thai & TH & 696 & 528 & 220 & 178 & 274 & 146 \\
Turkish & TR & 696 & 732 & 220 & 248 & 274 & 187   \\\bottomrule
\end{tabular}
}
\caption{Statistics of \lareqa{} in each 5-fold cross-validation split. \#Q denotes the number of question whereas \#C denotes the number of candidates.}
\label{tab:lareqa-stats}
\end{table}
\begin{table}[ht!]
\centering
\scalebox{0.95}{
\small
\begin{tabular}{l|l|lll}
\toprule
\multirow{2}{*}{\textbf{Language Pair}} & \multirow{2}{*}{\textbf{ISO}} & \multicolumn{3}{c}{\textbf{\# Sentence Pairs}} \\
& & Train & Dev & Test \\
\toprule
English-English  & EN-EN  & 150 & 50 & 50 \\
Spanish-Spanish  & ES-ES & 150 & 50 & 50 \\
Spanish-English  & ES-EN & 150 & 50 & 50 \\
Arabic-Arabic & AR-AR & 150 & 50 & 50 \\
Arabic-English & AR-EN  & 150 & 50 & 50 \\
Turkish-English* & TR-EN  & 150 & 50 & 50 \\
\bottomrule
\end{tabular}
}
\caption{\label{tab:stsbmulti-stats} Statistics of the \stsbm{} from SEM-Eval2007 in each 5-fold cross-validation split. * means that for Turkish-English, there are only 250 ground truth similarity scores, while there are 500 sentence pairs. We assume that the ground truth scores are only for the first 250 sentence pairs. In addition to that, we use 5749 train, 1500 dev, and 1379 test splits from the STSB original English benchmark.} 

%We use \num{150} for train, \num{50} for dev, and \num{50} for test in \stsbm{} from SEM-Eval2007 in each 5-fold cross-validation split. In addition to that, we use \num{5749} train, \num{1500} dev, and \num{1379} test splits from the STSB original English benchmark machine translated \meryem{specify how MT is used here and how this data is obtained} to different languages.
\end{table}

\subsection{Upstream Meta-Tasks}
\label{app:meta-tasks}

We detail in Table~\ref{tab:meta-tasks} the arrangements of languages for the different meta-tasks used in the meta-training $\metatraindata{}$, meta-validation $\metavaliddata{}$, and meta-testing $\metatestdata{}$ datasets. To make the comparison fair and consistent across different transfer modes, we use the same combination of languages and tweak them to fit the transfer mode. By picking a high number of meta-tasks during meta-training, meta-validation, and meta-testing, we make sure that all transfer modes are exposed to the same number of questions and candidates.
\begin{table*}[ht]
\centering
\scalebox{0.78}{
\begin{tabular}{l|l|l|l}
\toprule
\multirow{2}{*}{\textbf{Transfer Mode}} & \multirow{2}{*}{\textbf{Phase}} & \multicolumn{2}{c}{\textbf{Support$\rightarrow$Query/Support1$\rightarrow$Support2$\rightarrow$Query}} \\
& & \lareqa{} & \stsbm{} \\

\toprule

% \multirow{2}{*}{\begin{tabular}[c] \end{tabular}

% \multirow{2}{*}{\begin{tabular}[c]{@{}l@{}} \\ \end{tabular}}

\multirow{2}{*}{\monomono{}}  & \multirow{2}{*}{All}  &  \multirow{2}{*}{\begin{tabular}[c]{@{}l@{}} EL\_EL$\rightarrow$AR\_AR \\ HI\_HI$\rightarrow$DE\_DE \end{tabular}}   &\multirow{2}{*}{(EN\_EN,AR\_AR,ES\_ES)$\rightarrow$(EN\_EN,AR\_AR,ES\_ES)}  \\

& & & \\ 

% & Meta-valid  & \multirow{2}{*}{\begin{tabular}[c]{@{}l@{}} EL\_EL$\rightarrow$AR\_AR \\ HI\_HI$\rightarrow$DE\_DE \end{tabular}}  &  (EN\_EN,AR\_AR,ES\_ES)$\rightarrow$(EN\_EN,AR\_AR,ES\_ES)  \\

% & & & \\
%  & Meta-test  & \multirow{2}{*}{\begin{tabular}[c]{@{}l@{}} EL\_EL$\rightarrow$AR\_AR \\ HI\_HI$\rightarrow$DE\_DE \end{tabular}}    & (EN\_EN,AR\_AR,ES\_ES)$\rightarrow$(EN\_EN,AR\_AR,ES\_ES)   \\

% & & & \\
 
\midrule

\multirow{2}{*}{\monobil{}}  

& \multirow{2}{*}{All}  & \multirow{2}{*}{\begin{tabular}[c]{@{}l@{}} EL\_EL$\rightarrow$EL\_AR  \\ HI\_HI$\rightarrow$HI\_DE \end{tabular}}   & \multirow{2}{*}{[EN\_EN,AR\_AR,ES\_ES]$\rightarrow$[AR\_EN,ES\_EN,TR\_EN]} \\
 & & & \\
%& Meta-valid  & EL\_EL$\rightarrow$EL\_AR  \& HI\_HI$\rightarrow$HI\_DE & [EN\_EN,AR\_AR,ES\_ES]$\rightarrow$[AR\_EN,ES\_EN,TR\_EN] \\
 % & Meta-test  &  EL\_EL$\rightarrow$EL\_AR  \& HI\_HI$\rightarrow$HI\_DE & [EN\_EN,AR\_AR,ES\_ES]$\rightarrow$[AR\_EN,ES\_EN,TR\_EN] \\

\midrule

\multirow{2}{*}{\monomulti{}}
& \multirow{2}{*}{All}  &  
\multirow{2}{*}{\begin{tabular}[c]{@{}l@{}} EL\_EL$\rightarrow$EL\_\{AR,EL\} \\ HI\_HI$\rightarrow$HI\_\{DE,HI\} \end{tabular}} & \multirow{2}{*}{Not Applicable}  \\
& & & \\
 % & Meta-valid  &  EL\_EL$\rightarrow$EL\_{AR,EL}  \& HI\_HI$\rightarrow$HI\_{DE,HI}&\\
 % & Meta-test  & EL\_EL$\rightarrow$EL\_{AR,EL}  \& HI\_HI$\rightarrow$HI\_{DE,HI}\\

\midrule

\multirow{2}{*}{\bilmulti{}}  
& \multirow{2}{*}{All}  & \multirow{2}{*}{\begin{tabular}[c]{@{}l@{}} EL\_AR$\rightarrow$EL\_\{AR,EL\} \\  HI\_DE$\rightarrow$HI\_\{DE,HI\} \end{tabular}} & \multirow{2}{*}{Not Applicable}  \\
 % & Meta-valid  & EL\_AR$\rightarrow$EL\_{AR,EL} \&   HI\_DE$\rightarrow$HI\_{DE,HI}& \\
 % & Meta-test  &  EL\_AR$\rightarrow$EL\_{AR,EL} \&   HI\_DE$\rightarrow$HI\_{DE,HI}  \\

& & & \\
\midrule

\multirow{4}{*}{\mixt{}} 

& \multirow{4}{*}{All}  & 
\multirow{4}{*}{\begin{tabular}[c]{@{}l@{}} \monomono{} \\ \monobil{} \\ \monomulti{} \\ \bilmulti{} \end{tabular}}
%All of the above 
& \multirow{4}{*}{Not Applicable} \\
% & Meta-valid  & \monomono{} \& \monobil{} \& \monomulti{} \& \bilmulti{}  \\
% All of the above \\
%  & Meta-test  &  \monomono{} \& \monobil{} \& \monomulti{} \& \bilmulti{}  \\
 % All of the above \\

& & &  \\
& & &  \\
& & &  \\

\midrule

\multirow{3}{*}{\trans{}}  
& Meta-train  &  \monobil{} & \multirow{3}{*}{Not Applicable} \\
  & Meta-valid  &  \bilmulti{}  \\
  & Meta-test  & \monomulti{} & \\

\midrule

% \multirow{3}{*}{\begin{tabular}[c]{@{}l@{}}\textbf{Train} \\ \textbf{Language} \end{tabular}}

% \multirow{3}{*}{\begin{tabular}[c]{@{}l@{}}  \\  \end{tabular}}

\multirow{3}{*}{\monobilmulti{}}  & \multirow{3}{*}{All}  & \multirow{3}{*}{\begin{tabular}[c]{@{}l@{}}  EL\_EL$\rightarrow$EL\_AR$\rightarrow$EL\_\{AR,EL,HI\} \\  HI\_HI$\rightarrow$HI\_DE$\rightarrow$HI\_\{AR,DE,HI\} \end{tabular}}  &  \multirow{3}{*}{\begin{tabular}[c]{@{}l@{}} EN\_EN$\rightarrow$AR\_EN$\rightarrow$EN\_\{AR,EN,ES\}  \\ AR\_AR$\rightarrow$AR\_ES$\rightarrow$AR\_\{AR,EN,ES\} \\  ES\_ES$\rightarrow$ES\_AR$\rightarrow$ES\_\{AR,EN,ES\} \end{tabular}} \\
& & & \\

& & & \\

\bottomrule
\end{tabular}
}
\caption{Arrangements of languages for the different modes of transfer and meta-learning stages for two standard benchmark datasets \lareqa{} and \stsbm{}. X$\rightarrow$Y denotes transfer from an X model (for example a monolingual model) used to sample the support set to a Y model (for example bilingual model) used to sample the query set. We denote a support or query set in \lareqa{} by x\_y where x and y are the ISO language codes of the question and the candidate answers and x\_y in \stsbm{} where x and y are the ISO language codes of sentence 1 and 2 respectively. We use parenthesis to mean that the same language pairs cannot be used in both support and query sets, brackets to denote non-exclusivity (or in other words the language pairs used as a support can also be used as a query), and curled braces to mean the query set may be sampled from more than one language. We do not experiment with \monomulti{}, \bilmulti{}, \mixt{}, and \trans{} for \stsbm{}, since it is not a multilingual parallel benchmark, but we still experiment with \monobilmulti{} using machine-translated data in that case.}
\label{tab:meta-tasks}
\end{table*}

\subsection{Hyperparameters}
\label{app:hyperparam}

\begin{table}[ht!]
\centering
\scalebox{0.95}{
\small
\begin{tabular}{l|l}
\toprule
\textbf{Sentence Transformers Model} & \textbf{\map{}}  \\
\toprule
\laser{} & 13.5 \small{$\pm$ 0.7} \\
\labse{} &  48.7 \small{$\pm$ 2.6} \\
\mbert{}+\ensquad{} & 37.9 \small{$\pm$ 3.4}  \\
distilbert-multilingual-nli-stsb-quora-ranking & 44.1 \small{$\pm$ 0.9} \\
use-cmlm-multilingual & 36.8 \small{$\pm$ 2.6} \\
distiluse-base-multilingual-cased-v2 &  46.9 \small{$\pm$ 2.5} \\
paraphrase-multilingual-MiniLM-L12-v2 & 49.6  \small{$\pm$ 2.7} \\
multi-qa-distilbert-dot-v1 &  6.4 \small{$\pm$ 0.3} \\
paraphrase-multilingual-mpnet-base-v2 & \textbf{57.0} \small{$\pm$ 2.9} \\
\bottomrule
\end{tabular}
}
\caption{Comparison of \map{} multilingual 5-fold cross-validation evaluation of different \sbert{} models compared to \mbert{} model. Best results are highlighted in \textbf{bold}.}
\label{tab:baseline-models-lareqa}
\end{table}

Based on our prior investigation of different sentence-transformer models~in Table~\ref{tab:baseline-models-lareqa}, we notice that \textit{paraphrase-multilingual-mpnet-base-v2}\footnote{\url{https://huggingface.co/sentence-transformers/paraphrase-multilingual-mpnet-base-v2}.}, which maps sentences and paragraphs to a 768-dimensional dense vector space, performs the best for \lareqa{}, so we use it in our \sbert{} experiments on that dataset. The good initial performance of this pre-trained model is not surprising since it was trained on parallel data and is recommended for use in tasks like clustering or semantic search. For pre-processing \lareqa{} and \ensquad{}, we truncate/pad all questions to length 96 and all answer or negative candidates concatenated with their contexts to 256. For pre-processing \stsbm{} and \stsben{}, we pad or truncate each sentence to fit the maximum length of 100. 

For both benchmarks, for \finetune{} baselines, following XTREME-R, we use AdamW optimizer ~\cite{DBLP:conf/iclr/LoshchilovH19}. We use a learning rate of $lr=5e-5$, $\epsilon=1e-8$ and a weight decay of 0, with no decay on the bias and LayerNorm weights. We use a batch size of 8 triplets or sentence pairs. For \lareqa{}, we sample 3 negative examples per anchor and then project those to 3 triplets with one negative example and use a margin of 1. In \stsbm{}, we use just sets of sentence pairs composed of one source and one target sentence each, where we don't have negative examples so we don't need to flatten the dimensions of the negative examples. We sample 7,000, 2,000, and 1,000 meta-tasks in the meta-training, meta-validation, and meta-testing phases respectively. We use meta-batches of size $4$. In each meta-task, we randomly sample $k=8$ and $q=4$ support and query triplets respectively. We use the same meta-tasks and sampling regime in \finetune{} as well.

For \maml{} and \alignmaml{} in both benchmarks, we use learn2learn~\cite{arnold2020learn2learn} implementation to handle gradient updates, especially in the inner loop. For the inner loop, we use learn2learn pre-built optimizer with a learning rate $\alpha=1e-3$. The inner loop is repeated $n=5$ times for meta-training and meta-validation and meta-testing. For the outer loop, we use the same optimizer with the same learning rate $\beta=1e-5$ that we used in the \finetune{} model. At the end of each epoch, we perform meta-validation similarly to meta-training with the same hyperparameters described before. We use the same hyperparameters for \alignmaml{} for both \teachmaml{} and \studmaml{} except that we run the gradient updates in the inner loop in \studmaml{} just once, whereas for \teachmaml{} we perform $n=5$ inner loop gradient updates. We jointly optimize the outer loop losses weighting the knowledge distillation by $\lambda=0.5$. We don't use meta-testing but keep it for evaluation purposes. For a consistent comparison, we don't use meta-testing for our main evaluation as we use standard testing cross-validation splits, but we will include those meta-testing datasets to encourage future work on few-shot learning. All experiments are run for one fixed initialization seed using a 5-fold cross-validation. We observe a variance with respect to different seeds smaller than the variance with respect to 5-fold cross-validation, so we report the latter to have a better upper bound of the variance.

All experiments are conducted on the same computing infrastructure using \emph{one} NVIDIA A40 GPU with $46068$ MiB memory and \emph{one} TESLA P100-PCIE with $16384$ MiB memory of CUDA version 11.6 each. We use Pytorch version 1.11.1, Python version 3.8.13, learn2learn version 0.1.7, Hugging Face transformers version 4.21.3 and Sentence-Transformers 2.2.2. For paraphrase-multilingual-mpnet-base-v2 used in the experiments in the main paper, there are 278,043,648 parameters. For asymmetric and symmetric semantic search benchmarks, there are three and two encoding towers, respectively. Therefore, there are 
834,130,944 and 556,087,296 parameters used for asymmetric and symmetric semantic search benchmarks, respectively. 

For all experiments and model variants, we train for up to 20 epochs maximum and we implement early stopping, where we run the experiment for as long as there is an improvement on the \dev{} set performance. After 50 mini meta-task batches of no improvement on the \dev{} set, the experiment stops running. We use the multilingual performance on the \dev{} set averaged over all languages of the query set as the early stopping evaluation criteria. Based on this early stopping policy, we report in Table~\ref{tab:runtime} the typical runtime for each upstream model variant and baseline.

\begin{table}[ht!]
\centering
\scalebox{0.85}{
\begin{tabular}{lll}
\toprule
\textbf{Model} & \textbf{Runtime} \\ \toprule
\finetune{} &  2 h 18 min   \\
\maml{} &  3 h 19 min      \\
\alignmaml{} & 19 h 29 min       \\
  \bottomrule
\end{tabular}
}
\caption{Runtime per model variant excluding evaluation.}
\label{tab:runtime}
\end{table}

\section{More Results}
\label{app:more-results}

\begin{table*}[t] 
\centering
\scalebox{0.68}{ 
\begin{tabular}{l|l|llll|lll|l} \toprule
\multirow{4}{*}{\textbf{Model}} & \multirow{4}{*}{\begin{tabular}[c]{@{}l@{}}\textbf{Train Language(s)} \\ \textbf{Configuration} \end{tabular}} &
\multicolumn{8}{c}{\textbf{Testing Languages}}  \\
& & \multicolumn{4}{c|}{\textbf{Few-Shot Languages}} & \multicolumn{3}{c|}{\textbf{Zero-Shot Languages}} &   \\
& & \multicolumn{1}{c}{Arabic} &  \multicolumn{1}{c}{German} &  \multicolumn{1}{c}{Greek} & \multicolumn{1}{c|}{Hindi} & \multicolumn{1}{c}{Russian} & \multicolumn{1}{c}{Thai} & \multicolumn{1}{c|}{Turkish} & \multirow{2}{*}{Mean} \\ 
& & \multicolumn{1}{c}{AR} & \multicolumn{1}{c}{DE} & \multicolumn{1}{c}{EL} & \multicolumn{1}{c|}{HI} & \multicolumn{1}{c}{RU} & \multicolumn{1}{c}{TH} & \multicolumn{1}{c|}{TR} \\
\midrule
\rowcolor{lightgray}  \multicolumn{10}{c}{Zero-Shot Baselines}   \\ \hline
\laser{} & - & 13.2 \small{$\pm$ 5.1} & 15.1 \small{$\pm$ 5.9} & 14.6 \small{$\pm$ 5.6} & 9.4 \small{$\pm$ 3.9} & 14.9 \small{$\pm$ 5.6} & 13.0 \small{$\pm$ 5.6} & 14.1 \small{$\pm$ 6.0} & 13.5 \small{$\pm$ 0.7} \\
\labse{} & - & 44.7 \small{$\pm$ 2.0}
& 47.9 \small{$\pm$ 3.8}
& 53.0 \small{$\pm$ 3.4}
& 53.4 \small{$\pm$ 3.2}
& 53.1 \small{$\pm$ 3.8}
& 49.8 \small{$\pm$ 2.8}
& 48.1 \small{$\pm$ 3.5}
& 50.0 \small{$\pm$ 2.8} \\
\sbert{} & - & \underline{56.3} \small{$\pm$ 2.7} & 
\underline{54.6} \small{$\pm$ 2.1} &
\underline{58.2} \small{$\pm$ 3.8} &
\underline{57.2} \small{$\pm$ 3.9} &
\underline{58.7} \small{$\pm$ 3.3} &
\underline{60.2} \small{$\pm$ 3.4} &
\underline{54.1} \small{$\pm$ 2.5} & \underline{57.0} \small{$\pm$ 2.9}  \\
\rowcolor{lightgray}  \multicolumn{10}{c}{+Few-Shot Learning}   \\ \hline
%\cline{2-10}
\multirow{6}{*}{\sbert{}+\finetune{}}
& \monomono{} & \underline{45.9} \small{$\pm$ 2.4} & 46.3 \small{$\pm$ 2.5} & 47.9 \small{$\pm$ 2.6} & 45.4 \small{$\pm$ 3.1} & \underline{48.9} \small{$\pm$ 2.7} & \underline{49.7} \small{$\pm$ 3.2} & \underline{45.1} \small{$\pm$ 1.6} &  \underline{47.0} \small{$\pm$ 2.0} \\

& \monobil{} & 45.8 \small{$\pm$ 4.0} & \underline{46.5} \small{$\pm$ 3.6} & \underline{48.6} \small{$\pm$ 4.7} & \underline{45.0} \small{$\pm$ 5.8} & \underline{48.9} \small{$\pm$ 4.2} & 49.4 \small{$\pm$ 4.5} & 45.0 \small{$\pm$ 3.1} &  \underline{47.0} \small{$\pm$ 4.2} \\
& \monomulti{} & 40.4 \small{$\pm$ 3.9} & 42.5 \small{$\pm$ 3.2} & 43.1 \small{$\pm$ 4.6} & 37.8 \small{$\pm$ 5.4} & 44.1 \small{$\pm$ 4.3} & 44.3 \small{$\pm$ 4.4} & 41.1 \small{$\pm$ 3.1} &  41.9 \small{$\pm$ 4.0} \\
& \bilmulti{} & 33.8 \small{$\pm$ 4.9} & 35.6 \small{$\pm$ 4.2} & 35.2 \small{$\pm$ 6.2} & 32.4 \small{$\pm$ 3.9} & 37.1 \small{$\pm$ 5.3} & 37.2 \small{$\pm$ 5.5} & 34.4 \small{$\pm$ 4.3} &  35.1 \small{$\pm$ 4.8} \\
& \mixt{} & 38.3 \small{$\pm$ 4.1} & 39.8 \small{$\pm$ 4.6} & 40.7 \small{$\pm$ 3.8} & 39.3 \small{$\pm$ 5.2} & 41.9 \small{$\pm$ 5.0} & 41.7 \small{$\pm$ 5.1} & 38.7 \small{$\pm$ 3.9} &  40.1 \small{$\pm$ 4.4} \\
& \trans{} & 38.7 \small{$\pm$ 3.8} & 39.9 \small{$\pm$ 4.8} & 41.8 \small{$\pm$ 3.4} & 40.1 \small{$\pm$ 3.8} & 42.6 \small{$\pm$ 4.3} & 42.6 \small{$\pm$ 3.8} & 39.4 \small{$\pm$ 4.0} &  40.7 \small{$\pm$ 3.8} \\
\cline{2-10}
\multirow{6}{*}{\sbert{}+\maml{}} & \monomono{} & \underline{56.3} \small{$\pm$ 1.6} & 54.5 \small{$\pm$ 2.0} & 58.5 \small{$\pm$ 3.3} & \underline{57.0} \small{$\pm$ 2.5} & \underline{59.3} \small{$\pm$ 2.5} & 59.6 \small{$\pm$ 2.7} & 53.8 \small{$\pm$ 1.9} &  57.0 \small{$\pm$ 2.3} \\
& \monobil{} & 55.9 \small{$\pm$ 3.1} & \underline{55.0} \small{$\pm$ 3.0} & 58.4 \small{$\pm$ 4.6} & 56.9 \small{$\pm$ 4.0} & 58.8 \small{$\pm$ 3.9} & \underline{59.9} \small{$\pm$ 3.4} & 54.2 \small{$\pm$ 3.0} &  57.0 \small{$\pm$ 3.5} \\
& \monomulti{} & 54.9 \small{$\pm$ 2.8} & 53.6 \small{$\pm$ 3.4} & 57.0 \small{$\pm$ 4.7} & 55.8 \small{$\pm$ 3.9} & 57.7 \small{$\pm$ 4.1} & 58.7 \small{$\pm$ 3.4} & 53.1 \small{$\pm$ 3.2} &  55.9 \small{$\pm$ 3.5} \\
& \bilmulti{} & 54.5 \small{$\pm$ 2.1} & 53.6 \small{$\pm$ 2.1} & 56.6 \small{$\pm$ 2.5} & 55.5 \small{$\pm$ 1.7} & 57.3 \small{$\pm$ 2.2} & 58.5 \small{$\pm$ 1.9} & 52.8 \small{$\pm$ 1.6} &  55.5 \small{$\pm$ 1.7} \\
& \mixt{} & 55.0 \small{$\pm$ 3.1} & 53.9 \small{$\pm$ 2.4} & 57.2 \small{$\pm$ 3.9} & 55.3 \small{$\pm$ 4.2} & 57.6 \small{$\pm$ 3.7} & 58.7 \small{$\pm$ 3.0} & 52.9 \small{$\pm$ 3.0} &  55.8 \small{$\pm$ 3.2} \\
& \trans{}
& 56.0 \small{$\pm$ 3.7} & 54.8 \small{$\pm$ 2.2} & \underline{59.1} \small{$\pm$ 4.2} & \underline{57.0} \small{$\pm$ 4.4} & 59.1 \small{$\pm$ 4.1} & \underline{59.9} \small{$\pm$ 3.8} & \underline{54.4}  \small{$\pm$ 3.0} & \underline{57.2} \small{$\pm$ 3.5}   \\ \cline{2-10}
\sbert{}+\alignmaml{} & \monobilmulti{} 
& \textit{\underline{57.0}} \small{$\pm$ 2.9} &
\textit{\underline{55.1}} \small{$\pm$ 2.4} & 
\textit{\underline{59.2}} \small{$\pm$ 4.2} & 
\textit{\underline{57.7}} \small{$\pm$ 4.5} &
\textit{\underline{59.5}} \small{$\pm$ 3.5} &
\textit{\underline{60.2}} \small{$\pm$ 3.7} &
\textit{\underline{54.6}} \small{$\pm$ 2.7} & \textit{\underline{57.6}} \small{$\pm$ 3.3}  \\
 \rowcolor{lightgray}
\multicolumn{10}{c}{+Machine Translation}   \\ \hline
%\sbert{}+\translatetest{} & - & \textit{62.8} \small{$\pm$ 2.3} & \textit{64.4} \small{$\pm$ 2.3} & \textit{64.2} \small{$\pm$ 1.7} & \textit{63.4} \small{$\pm$ 1.4} & \textit{63.2} \small{$\pm$ 2.0} & \textit{62.4} \small{$\pm$ 2.6} & \textit{64.1} \small{$\pm$ 2.3} & \textit{63.5} \small{$\pm$ 0.4} \\
%\sbert{}+\maml{}+\translatetest{}& Best transfer mode & \textbf{\underline{63.7}} \small{$\pm$ 2.6} & \textbf{\underline{65.0}} \small{$\pm$ 2.6} & \textbf{\underline{64.8}} \small{$\pm$ 2.3} & \textbf{\underline{64.1}} \small{$\pm$ 2.0} & \textbf{\underline{64.0}} \small{$\pm$ 2.4} & \textbf{\underline{63.1}} \small{$\pm$ 2.5} & \textbf{\underline{64.8}} \small{$\pm$ 2.4} &  \textbf{\underline{64.2}} \small{$\pm$ 0.2}  \\ \hline{}
 \multirow{5}{*}{\sbert{}+\translatetrain{}+\finetune{}} & AR\_AR$\rightarrow$AR\_AR & \underline{46.6} \small{$\pm$ 3.5} & \underline{45.8} \small{$\pm$ 3.4} & \underline{48.8} \small{$\pm$ 4.2} & \underline{46.8} \small{$\pm$ 4.2} & \underline{49.3} \small{$\pm$ 4.6} & 48.6 \small{$\pm$ 3.8} & \underline{44.9} \small{$\pm$ 3.5} & \underline{47.3} \small{$\pm$ 3.8} \\
 & DE\_DE$\rightarrow$DE\_DE & 45.9 \small{$\pm$ 5.0} & 45.1 \small{$\pm$ 4.4} & 48.2 \small{$\pm$ 5.8} & 45.8 \small{$\pm$ 6.5} & 49.0 \small{$\pm$ 5.1} & \underline{48.8} \small{$\pm$ 6.8} & 44.5 \small{$\pm$ 4.5} & 46.8 \small{$\pm$ 5.4} \\
 & EL\_EL$\rightarrow$EL\_EL & 43.5 \small{$\pm$ 4.3} & 43.1 \small{$\pm$ 4.5} & 43.8 \small{$\pm$ 4.5} & 43.4 \small{$\pm$ 4.1} & 46.5 \small{$\pm$ 4.2} & 45.0 \small{$\pm$ 3.5} & 41.7 \small{$\pm$ 4.3} & 43.8 \small{$\pm$ 4.0} \\
 & HI\_HI$\rightarrow$HI\_HI & 46.5 \small{$\pm$ 3.1} & 44.8 \small{$\pm$ 2.9} & 47.1 \small{$\pm$ 3.8} & 45.9 \small{$\pm$ 4.1} & 48.4 \small{$\pm$ 4.4} & 49.6 \small{$\pm$ 3.7} & 43.7 \small{$\pm$ 3.0} & 46.6 \small{$\pm$ 3.4} \\
 & All test languages & 44.8 \small{$\pm$ 2.8} & 43.5 \small{$\pm$ 3.2} & 46.9 \small{$\pm$ 4.0} & 44.0 \small{$\pm$ 4.5} & 47.0 \small{$\pm$ 3.4} & 46.4 \small{$\pm$ 3.9} & 42.1 \small{$\pm$ 3.0} & 45.0 \small{$\pm$ 3.4} \\ \cline{2-10}
\multirow{5}{*}{\sbert{}+\translatetrain{}+\maml{}} & AR\_AR$\rightarrow$AR\_AR & \textbf{\underline{57.3}} \small{$\pm$ 3.2} & \textbf{\underline{55.3}} \small{$\pm$ 2.1} & \textbf{\underline{59.3}} \small{$\pm$ 4.2} & \textbf{\underline{58.3}} \small{$\pm$ 4.2} & \textbf{\underline{60.2}} \small{$\pm$ 3.6} & \textbf{\underline{60.7}} \small{$\pm$ 3.5} & \textbf{\underline{54.8}} \small{$\pm$ 2.3} & \textbf{\underline{58.0}} \small{$\pm$ 3.2} \\
& DE\_DE$\rightarrow$DE\_DE &  56.1 \small{$\pm$ 2.7} & 54.4 \small{$\pm$ 2.2} & 58.3 \small{$\pm$ 3.9} & 57.1 \small{$\pm$ 4.1} & 58.8 \small{$\pm$ 3.9} & 59.8 \small{$\pm$ 3.7} & 54.1 \small{$\pm$ 2.7} &  56.9 \small{$\pm$ 3.2} \\
& EL\_EL$\rightarrow$EL\_EL & 55.9 \small{$\pm$ 3.4} & 53.1 \small{$\pm$ 4.3} & 57.4 \small{$\pm$ 5.2} & 56.3 \small{$\pm$ 5.5} & 58.5 \small{$\pm$ 4.6} & 59.2 \small{$\pm$ 5.2} & 52.8 \small{$\pm$ 4.4} & 56.2 \small{$\pm$ 4.5} \\
& HI\_HI$\rightarrow$HI\_HI & 56.7 \small{$\pm$ 3.6} & 54.0 \small{$\pm$ 2.6} & 58.5 \small{$\pm$ 4.6} & 57.1 \small{$\pm$ 4.7} & 58.9 \small{$\pm$ 4.1} & 60.3 \small{$\pm$ 3.0} & 53.7 \small{$\pm$ 3.3} & 57.0 \small{$\pm$ 3.5} \\
& All test languages  & 55.9 \small{$\pm$  3.9} & 53.8 \small{$\pm$ 2.7} & 58.0 \small{$\pm$ 4.9} & 56.6 \small{$\pm$ 4.4} & 58.1 \small{$\pm$ 4.3} & 59.2 \small{$\pm$ 3.9} & 53.4 \small{$\pm$ 3.0} & 56.4 \small{$\pm$ 3.7} \\
\bottomrule
\end{tabular}
}
\caption{\label{tab:asym-search-lareqa-languages} \map{} multilingual 5-fold cross-validated performance tested for different languages. Best and second-best results for each language are highlighted in \textbf{bold} and \textit{italicized} respectively, whereas best results across categories of models are \underline{underlined}. Gains from meta-learning approaches are consistent across few-shot and zero-shot languages.}
\end{table*}

\begin{table*}[t!] 
\centering
\scalebox{0.58}{ 
\begin{tabular}{l|l|llllll|l} \toprule \multirow{3}{*}{\textbf{Model}} & \multirow{3}{*}{\begin{tabular}[c]{@{}l@{}}\textbf{Train Language(s)} \\ \textbf{Configuration} \end{tabular}}  &
\multicolumn{7}{c}{\textbf{Testing Languages}}  \\
& & Arabic-Arabic & Arabic-English & Spanish-Spanish & Spanish-English & English-English & Turkish-English & \multirow{2}{*}{Mean} \\ 
& & AR-AR & AR-EN & ES-ES & ES-EN & EN-EN & TR-EN \\
\rowcolor{lightgray}  \multicolumn{9}{c}{Zero-Shot Learning}   \\ \hline
\laser{} & - &  22.5 \small{$\pm$ 8.5}
& 21.6 \small{$\pm$ 8.4}
& 33.1 \small{$\pm$ 9.4}
& 15.3 \small{$\pm$ 15.7}
& 31.1 \small{$\pm$ 5.4}
& 21.2 \small{$\pm$ 13.7}
& 24.1 \small{$\pm$ 12.4} \\
\labse{} & - &  71.6 \small{$\pm$ 6.2}
& 73.2 \small{$\pm$ 4.0}
& 83.2 \small{$\pm$ 1.7}
& 68.7 \small{$\pm$ 10.1}
& 76.3 \small{$\pm$ 2.7}
& 74.9 \small{$\pm$ 3.3} & 74.6 \small{$\pm$ 4.6} \\
\sbert{} & - &   \underline{77.6} \small{$\pm$ 5.3}
 & \underline{81.3} \small{$\pm$ 3.2} 
 & \underline{84.6} \small{$\pm$ 2.9} 
 & \underline{83.7} \small{$\pm$ 6.7} 
 & \underline{85.5} \small{$\pm$ 4.2} 
 & \underline{75.7} \small{$\pm$ 3.1}  & \underline{81.4} \small{$\pm$ 4.2} \\
\rowcolor{lightgray}
\multicolumn{9}{c}{+Few-Shot learning}   \\ \hline
\sbert{}+\finetune{} & \monobil{} &   77.2 \small{$\pm$ 5.8}
& 77.8 \small{$\pm$ 3.8}
& 86.2 \small{$\pm$ 2.8}
& 79.6 \small{$\pm$ 8.3}
& 85.0 \small{$\pm$ 4.5}
& 73.7 \small{$\pm$ 4.6}
& 79.9 \small{$\pm$ 2.0} \\ % \finetune{} & {'ar-ar': (77.2, 5.8), 'es-es': (86.2, 2.8), 'en-en': (85.0, 4.5), 'ar-en': (77.8, 3.8), 'es-en': (79.6, 8.3), 'tr-en': (73.7, 4.6)} \\
% finetune{} one seed MEAN: 78.0 1.8  BIL: 74.4 3.0  MONO: 81.6 3.4 {'ar-ar': (76.3, 6.8), 'es-es': (84.5, 3.1), 'en-en': (83.8, 4.3), 'ar-en': (75.5, 4.3), 'es-en': (77.2, 8.8), 'tr-en': (70.6, 4.5)}
\sbert{}+\maml{} & \monobil{} & 
77.6 \small{$\pm$  5.3} & \underline{80.9} \small{$\pm$ 2.6} &   85.1 \small{$\pm$ 2.4} & \underline{83.5} \small{$\pm$ 6.7} & 85.6 \small{$\pm$ 4.8} & 75.5 \small{$\pm$ 3.7} & 81.3 \small{$\pm$ 1.4}  \\
% \sbert{}+\maml{} MEAN: 0.8133610484523695  BIL: 79.9 2.9  MONO: 82.7 3.2 { 77.6 \small{$\pm$  5.3} & 80.9 \small{$\pm$ 2.6} &   85.1 \small{$\pm$ 2.4} & 83.5 \small{$\pm$ 6.7} & 85.6 \small{$\pm$ 4.8} & 75.5 \small{$\pm$ 3.7} \\
\sbert{}+\alignmaml{} & \monobilmulti{} & \textbf{\underline{79.0}} \small{$\pm$ 5.2} & 80.6 \small{$\pm$ 1.0} & \underline{86.6} \small{$\pm$ 2.1} & 81.5 \small{$\pm$ 6.8} & \underline{\textbf{90.6}} \small{$\pm$ 1.1} & \textbf{\underline{76.3}} \small{$\pm$ 4.0} & \textbf{\underline{82.4}} \small{$\pm$ 1.4} \\
%MEAN: 82.4 1.4  BIL: 79.5 2.7  MONO: 85.4 1.3 ( 79.0 \small{$\pm$ 5.2} & 80.6 \small{$\pm$ 1.0} & 86.6 \small{$\pm$ 2.1} & 81.5 \small{$\pm$ 6.8} & 90.6 \small{$\pm$ 1.1} & 76.3 \small{$\pm$ 4.0} & 82.4 \small{$\pm$ 1.4} \\
\rowcolor{lightgray}
\multicolumn{9}{c}{+Machine Translation}   \\  \hline
%\sbert{}+\translatetest{} & - & \textit{80.7} \small{$\pm$ 4.1} & \textit{\underline{81.3}} \small{$\pm$ 3.2} & \textbf{\underline{88.6}} \small{$\pm$ 2.9} & \textbf{\underline{85.7}} \small{$\pm$ 6.2} & \underline{85.5} \small{$\pm$ 4.2} & \underline{75.7} \small{$\pm$ 3.1} & \textbf{\underline{82.9}} \small{$\pm$ 5.9}   \\
% \sbert{}+\maml{}+\translatetest{} & Best transfer mode & \textbf{\underline{81.3}} \small{$\pm$ 7.0} & 72.7 \small{$\pm$ 5.8} & 87.2 \small{$\pm$ 1.4} & 81.5 \small{$\pm$ 6.4} & 83.0 \small{$\pm$ 4.0} & 71.8 \small{$\pm$ 4.2} & 79.6 \small{$\pm$ 7.5} \\ \hline
\multirow{4}{*}{\sbert{}+\translatetrain{}+\finetune{}} 
 & AR\_AR$\rightarrow$AR\_AR & 59.5 \small{$\pm$ 7.9} & 50.6 \small{$\pm$ 12.5} & 82.7 \small{$\pm$ 4.3} & 70.1 \small{$\pm$ 11.9} &  82.4 \small{$\pm$ 5.7} & 62.5 \small{$\pm$ 5.9} & 68.0 \small{$\pm$ 14.6} \\ % bil 61.1 13.2 mono 74.9 12.5

  & EN\_EN$\rightarrow$EN\_EN & 72.6 \small{$\pm$ 6.8} & 73.1 \small{$\pm$ 4.9} & 82.4 \small{$\pm$ 2.9} & 72.2 \small{$\pm$ 10.9} & 80.3 \small{$\pm$ 6.8} & \underline{68.8} \small{$\pm$ 5.8} & 74.9 \small{$\pm$ 8.3}\\ %71.4 7.9 78.4 7.2
 & ES\_ES$\rightarrow$ES\_ES & \underline{74.2} \small{$\pm$ 8.0} & 72.3 \small{$\pm$ 8.0} & 82.3 \small{$\pm$ 2.8} & 66.8 \small{$\pm$ 12.1} & 79.7 \small{$\pm$ 6.9} & 68.5 \small{$\pm$ 4.8} & 73.9 \small{$\pm$ 9.5}  \\ % 69.2 9.1 78.7 7.2

  & TR\_TR$\rightarrow$TR\_TR & 73.9 \small{$\pm$ 6.3} & \underline{74.6} \small{$\pm$ 3.4} & \underline{85.9} \small{$\pm$ 2.0} & \underline{79.6} \small{$\pm$ 6.3} & \underline{84.3} \small{$\pm$ 4.7} & 68.5 \small{$\pm$ 3.7} &
\underline{77.8} \small{$\pm$ 7.7}  \\ % 74.3 6.5 81.4 7.1

 & All test languages & 65.8 \small{$\pm$ 9.0} & 63.0 \small{$\pm$ 4.6} & 82.5 \small{$\pm$ 3.0} & 75.8 \small{$\pm$ 8.7} & 83.0 \small{$\pm$ 4.7} & 67.8 \small{$\pm$ 4.9} & 73.0 \small{$\pm$ 10.1} \\ \cline{2-9} % 68.9 8.3 77.1 10.1 
\multirow{4}{*}{\sbert{}+\translatetrain{}+\maml{}}
& AR\_AR$\rightarrow$AR\_AR & 75.5 \small{$\pm$ 6.0} & 80.5 \small{$\pm$  2.5} & 85.8 \small{$\pm$ 2.1} & 83.1 \small{$\pm$ 6.3} & 85.6 \small{$\pm$ 3.9} & 75.0 \small{$\pm$ 4.0}
& 80.9 \small{$\pm$ 6.2}  \\ % 79.5 5.7 82.3 6.4

& EN\_EN$\rightarrow$EN\_EN & \textit{\underline{77.8}} \small{$\pm$ 5.2} & \textit{81.7} \small{$\pm$ 3.0} & 85.1 \small{$\pm$ 2.6} & \textbf{\underline{83.8}} \small{$\pm$ 6.6} & \textit{\underline{85.7}} \small{$\pm$ 4.3} & 75.8 \small{$\pm$ 3.5} &
\textit{\underline{81.6}} \small{$\pm$ 5.8} \\   % 80.4 5.7 82.8 5.5

& ES\_ES$\rightarrow$ES\_ES & 76.4 \small{$\pm$ 6.4} & 79.4 \small{$\pm$ 3.4} & \textit{86.9} \small{$\pm$ 1.6} & 80.4 \small{$\pm$ 7.7} & 84.7 \small{$\pm$ 4.7} & 74.1 \small{$\pm$ 4.2} & 80.3 \small{$\pm$ 6.7} \\ % 78.0 6.1 82.7 6.5

& TR\_TR$\rightarrow$TR\_TR & 77.2 \small{$\pm$ 5.9} & 79.8 \small{$\pm$ 3.8} & \textbf{\underline{87.3}} \small{$\pm$ 1.7} & 81.6 \small{$\pm$ 6.4} & 84.5 \small{$\pm$ 4.2} & 74.2 \small{$\pm$ 2.2} & 80.8 \small{$\pm$ 6.2} \\ % 78.5 5.5 83.0 6.0

& All test languages &  77.6 \small{$\pm$ 5.3} & \textbf{\underline{81.8}} \small{$\pm$ 2.5} & 84.7 \small{$\pm$ 2.9} & 83.6 \small{$\pm$ 6.7} & 85.6 \small{$\pm$ 4.2} & \textit{\underline{75.9}} \small{$\pm$ 3.4} & 
81.5 \small{$\pm$ 5.7} \\

\bottomrule
\end{tabular}
}
\caption{\label{tab:sym-search-stsb-languages} Pearson correlation \pearcorr{} 5-fold cross-validated performance on \stsbm{} benchmark using different models few-shot learned on \stsbm{} or its translation. Best and second-best results for each language are highlighted in \textbf{bold} and \textit{italicized} respectively, whereas best results across categories of models are \underline{underlined}.}
\end{table*}

Tables~\ref{tab:asym-search-lareqa-languages} and~\ref{tab:sym-search-stsb-languages} show full fine-grained results for all languages and language pairs for both semantic search benchmarks.

\end{document}